\documentclass[preprint,12pt]{elsarticle}
\usepackage[top=1in,bottom=1in,left=1in,right=1in]{geometry}
\usepackage{graphicx}
\usepackage{epstopdf, epsfig}
\usepackage{amsmath}

\usepackage{hhline}
\usepackage{tikz}
\usepackage{amssymb}
\usepackage{color}
\usepackage{float}
\usepackage[english]{babel}
\usepackage{tabularx}
\usepackage{algpseudocode}
\usepackage{algorithm}
\usepackage{hyperref}
\usepackage{bm}

\usepackage{graphicx}
\usepackage{listings}
\usepackage{subfigure}
\usepackage{float}
\def\temp {T}

\DeclareMathOperator*{\argmin}{arg\,min}

\biboptions{sort&compress}

\begin{document}

\begin{frontmatter}

\title{NeuroSEM: A hybrid framework for simulating multiphysics problems by coupling PINNs and spectral elements}

\author[brown]{Khemraj Shukla \fnref{1}}
\author[brown]{Zongren Zou \fnref{1}}
\author[imperial]{Chi Hin Chan}
\author[brown]{Additi Pandey}
\author[brown]{Zhicheng Wang}
\author[brown,brown2,pnnl]{George Em Karniadakis\fnref{2}}
\date{}

\address[brown]{Division of Applied Mathematics, Brown University, Providence, RI 02906, USA}
\address[imperial]{Department of Aeronautics, Imperial College London, South Kensington, London SW7 2AZ, UK}
\address[brown2]{School of Engineering, Brown University, Providence, RI 02906, USA}
\address[pnnl]{Pacific Northwest National Laboratory, Richland, WA 99354, USA}

\fntext[1]{These authors contributed equally to this work.}
\fntext[2]{Corresponding author: George Em Karniadakis (george\_karniadakis@brown.edu).}

\begin{abstract}
Multiphysics problems that are characterized by complex interactions among fluid dynamics, heat transfer, structural mechanics, and electromagnetics, are inherently challenging due to their coupled nature. While experimental data on certain state variables may be available, integrating these data with numerical solvers remains a significant challenge. Physics-informed neural networks (PINNs) have shown promising results in various engineering disciplines, particularly in handling noisy data and solving inverse problems in partial differential equations (PDEs). However, their effectiveness in forecasting nonlinear phenomena in multiphysics regimes, particularly involving turbulence, is yet to be fully established.

This study introduces NeuroSEM, a hybrid framework integrating PINNs with the high-fidelity Spectral Element Method (SEM) solver, Nektar++. NeuroSEM leverages the strengths of both PINNs and SEM, providing robust solutions for multiphysics problems. PINNs are trained to assimilate data and model physical phenomena in specific subdomains, which are then integrated into the Nektar++ solver. We demonstrate the efficiency and accuracy of NeuroSEM for thermal convection in cavity flow and flow past a cylinder. The framework effectively handles data assimilation by addressing those subdomains and state variables where the data is available.

We applied NeuroSEM to the Rayleigh-Bénard convection system, including cases with missing thermal boundary conditions and noisy datasets. {\color{red}Finally, we applied the proposed NeuroSEM framework to real particle image velocimetry (PIV) data to capture flow patterns characterized by horseshoe vortical structures.} Our results indicate that NeuroSEM accurately models the physical phenomena and assimilates the data within the specified subdomains. The framework's plug-and-play nature facilitates its extension to other multiphysics or multiscale problems. Furthermore, NeuroSEM is optimized for efficient execution on emerging integrated GPU-CPU architectures. This hybrid approach enhances the accuracy and efficiency of simulations, making it a powerful tool for tackling complex engineering challenges in various scientific domains.
\end{abstract}

\begin{keyword}
physics-informed machine learning, PINNs, spectral element method, data assimilation, PIV, multiphysics problems, heat transfer, domain decomposition 
\end{keyword}

\end{frontmatter}

\section{Introduction}
Data assimilation has been routinely employed in geophysics and weather forecasting \cite{law2015data, fletcher2022data, evensen2022data, gelfand2010handbook, carrassi2018data, reichle2008data, bouttier2002data, lahoz2010data, asch2016data} but is not as often utilized in engineering applications, e.g., in multiphysics problems such as mixed heat convection, magneto-hydrodynamics, hypersonics, etc., due to the increased volume of data available for engineering applications. Hence, it is important to develop computational methods that seamlessly integrate data into numerical simulations. However, for existing methods like finite element methods (FEM), which are typically used in complex industrial engineering problems, incorporating the data into FEM codes requires elaborate data assimilation techniques {\color{cyan}such as 3D-VAR, 4D-VAR and Kalman filters \cite{law2015data, fletcher2022data, evensen2022data}} that increase computational cost substantially. Physics-informed neural networks (PINNs) \cite{raissi2019physics}, first proposed in 2017 \cite{raissi2017physics,raissi2017physicsinformeddeeplearning}, enable seamless (at no extra cost) integration of multimodal data, which may be scattered measurements (e.g., distributed thermocouples) or images (e.g., infrared camera). PINNs can also make predictions like FEM without requiring any data in solving forward problems, but at the present time, they are not as accurate as the existing high-fidelity solvers, and are also comparatively more expensive.

\begin{figure}
\centering
\includegraphics[width=0.9\textwidth]{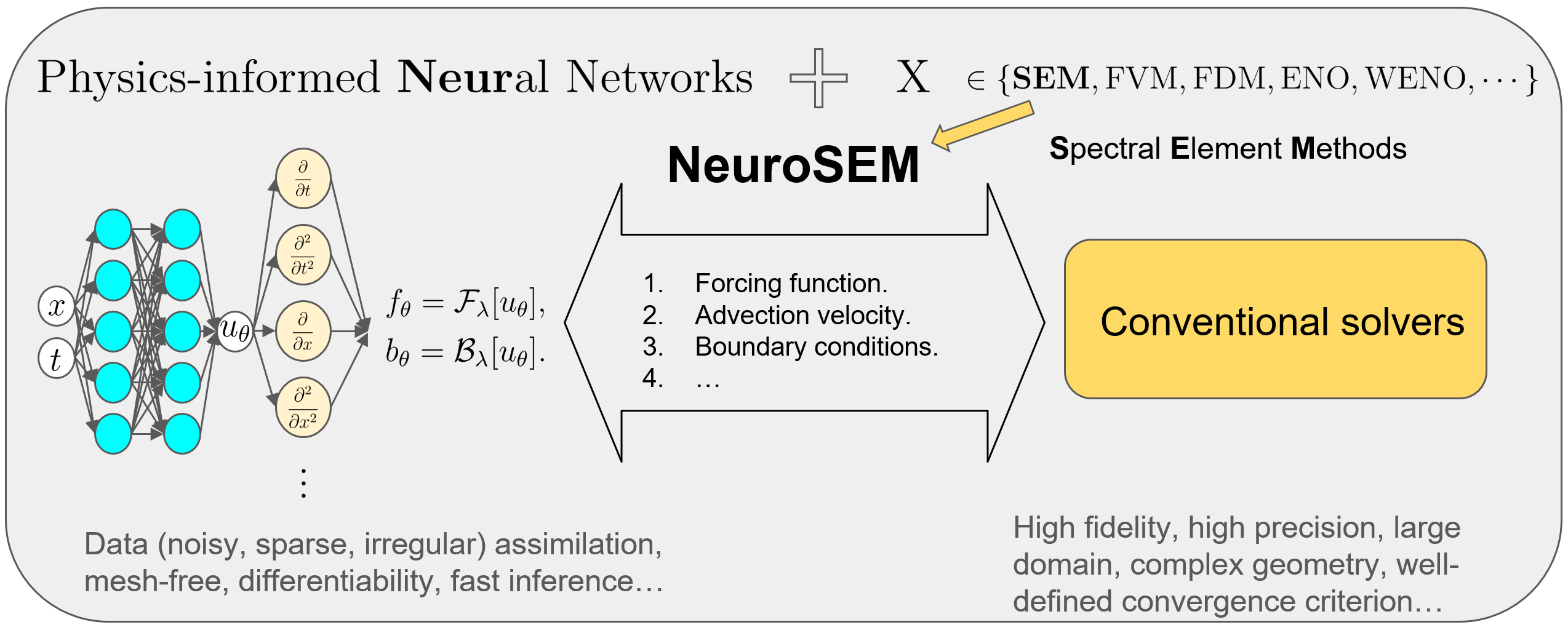}
\caption{Coupling $\text{PINNs}$ and $\text{X}$, where $\text{X}$ can be any existing numerical methods, e.g. SEM, FVM and FDM. In this study, we choose SEM as the backbone and propose the \textbf{NeuroSEM} method to solve the multiphysics problem.}
\label{fig:1}
\end{figure}

PINNs have proven to be very effective in solving severely ill-posed problem and have been successfully employed to incorporate the experimental data into PDEs. Examples include problems in fluid mechanics \cite{raissi2020hidden, almajid2022prediction, cheng2021deep, wessels2020neural, jin2021nsfnets, eivazi2022physics}, non-destructive evaluation of materials \cite{shukla2020physics, shukla2021physics}, gray box learning of model \cite{kiyani2023framework, zhang2024discovering, zou2023correcting}, and many more. Rigorous reviews of the work related to PINNs have been presented in \cite{karniadakis2021physics, cuomo2022scientific}.
In this work, we leverage the advantages of {\em both} PINNs and classical numerical methods by integrating them into a {\em unified} framework, called NeuroSEM, that can scale up to any size of multiphysics or multiscale problems by using standard domain decomposition techniques. 
As an example of existing numerical methods, we employ the high-order spectral element (SEM) solver Nektar++ \cite{cantwell2015nektar++}, but any other solver, e.g., OpenFOAM \cite{jasak2009openfoam} can also be employed. The main idea is to limit the use of PINNs only in those subdomains and state variables for which data is available, and employ SEM in the rest of the domain. Coupling is done via Dirichlet boundary conditions as we demonstrate in various scenarios that we consider herein but this coupling can be extended to impose proper interface conditions, including fluxes, stresses, etc., using standard domain decomposition methods, such as discontinuous Galerkin \cite{cockburn2012discontinuous} or 
the conservative PINNs \cite{jagtap2020conservative}. A schematic view of PINNs $+$ X where X can be any existing high-fidelity solver is presented in Fig. \ref{fig:1}. This coupling is simplified in NeuroSEM because in the data subdomains, PINNs can infer the interface boundary conditions, e.g., see our previous work on hidden fluid mechanics \cite{raissi2020hidden} or the work on missing thermal boundary conditions \cite{cai2021physics}. The rest of the domain, {\color{cyan}however complex its geometry is}, can be covered by spectral elements, leading to a high-accuracy forecasting. This type of coupling avoids costly iterations and is amenable to parallelization and easy implementation in the emerging integrated GPU-CPU computer architectures.

\textbf{Prototype problem:} As a proof of concept, we consider the Rayleigh-B\'enard convection described by the following form \cite{busse1978non} defined on the spatial-temporal domain $\Omega \times [0, \mathcal{T}]$:
\begin{subequations}\label{eq:problem}
    \begin{align}
        \nabla \cdot \bm{u} &= 0 \label{eq:problem_a},\\
        \frac{\partial \bm{u}}{\partial t} + (\bm{u} \cdot \nabla)\bm{u} &= -\nabla p + \frac{1}{\text{Re}} \nabla ^2 \bm{u} + \text{Ri} \temp\bm{k} \label{eq:problem_b},\\
        \frac{\partial \temp}{\partial t} + \bm{u} \cdot \nabla \temp &= \frac{1}{\text{Pe}}\nabla^2 \temp, \label{eq:problem_c}
    \end{align}
\end{subequations}
where $\bm{u} = [u, v]^T,~p,~\temp$ are the non-dimensional velocity, pressure and temperature, respectively, $\bm{k}$ is a unit vector, describing the direction of the gravity ($g$), and $\text{Re}$, $\text{Ri}$  and $\text{Pe}$ are Reynolds, Richardson and P\'eclet numbers, respectively.
The spatial and temporal domains, along with the associated initial and boundary conditions, vary depending on the specific problem.
In this study, we focus on two different problem setups: (1) steady-state flows on a square domain, and (2) an unsteady flow past a cylinder. We consider multiple scenarios, assuming that some data (probably noisy) are available for velocity $\bm{u}$ or temperature $\temp$ or both in the entire domain or in a small region, which could be a case of an experiment using PIV for velocity data or an infrared camera for the temperature data, where the spatial extent of data is rather limited. 

The paper is organized as follows: in Sec. \ref{sec:2}, we present the methodology and discuss in detail how we couple PINNs with SEM solvers to address the multiphysics problem; in Sec. \ref{sec:3}, we demonstrate the effectiveness of the proposed NeuroSEM method by employing it to solve \eqref{eq:problem} with different domains under various scenarios; we summarize in Sec. \ref{sec:4}.

\section{Methodology: NeuroSEM}\label{sec:2}

In this work, we consider a combination of the PINNs method and the spectral element method (SEM) as the backbone for the integration of machine learning and conventional numerical techniques. 

\subsection{Physics-informed neural networks (PINNs)}

The PINNs method \cite{raissi2019physics} leverages NNs and modern machine learning techniques to solve PDEs and has proven to be effective in addressing forward and inverse problems across various disciplines \cite{raissi2020hidden, almajid2022prediction, cheng2021deep, wessels2020neural, jin2021nsfnets, shukla2020physics, shukla2021physics, kiyani2023framework, zhang2024discovering, zou2023correcting, jagtap2020conservative, cai2021physics, cai2021physics_fluid, linka2022bayesian, henkes2022physics, zou2024neuraluq, mao2020physics, zou2023uncertainty, shukla2021parallel, psaros2023uncertainty, misyris2020physics, lu2021physics, zou2023hydra, haghighat2021physics, chen2024leveraging0, chen2024leveraging, sahli2020physics, eivazi2022physics, bin2021pinneik}.
In PINNs, the sought solution to \eqref{eq:problem} is modeled with NNs, denoted as $\bm{u}_\theta, p_\theta, \temp_\theta$ where $\theta$ is the NN parameter, and the physics-informed loss function is constructed via automatic differentiation (AD) \cite{raissi2019physics, lu2021deepxde} to encode the PDE information:
\begin{equation}\label{PINN_loss}
    \mathcal{L}(\theta) = w_{\mathcal{D}} \mathcal{L}_{\mathcal{D}}(\theta) + w_{PDE}\mathcal{L}_{PDE}(\theta),
\end{equation}
where $\mathcal{L}_{\mathcal{D}}$ and $\mathcal{L}_{PDE}$ are loss functions for data and PDE, respectively, and $w_{\mathcal{D}}, w_{PDE}$ are hyperparameters representing  belief weights. 
When PINNs are employed to solve \textit{inverse problems}, in which some of the PDE terms are unknown, these terms are modeled with additional NNs and then considered in  $\mathcal{L}_{PDE}$.
Certain optimization techniques such as stochastic gradient descent \cite{bottou2010large, kingma2014adam} are employed to obtain the minimizer $\theta^* = \argmin_\theta\mathcal{L}(\theta)$ and the approximated solution $\bm{u}_{\theta^*}, p_{\theta^*}, \temp_{\theta^*}$.
Despite being an effective method for solving severely ill-posed inverse problems, PINNs may lack high accuracy and efficiency in solving PDE systems on large domains, especially for forward problems, due to the following reasons: (1) the high cost of AD in computing high-order partial derivatives; (2) the large number of residual points required by the physics-informed loss function; and (3) the  difficulty of training NNs such that the global minimum of the loss function is obtained \cite{karniadakis2021physics, cuomo2022scientific, wang2021understanding, krishnapriyan2021characterizing}.




\subsection{Integration of PINNs with Nektar++}

Spectral element methods (SEM) are high-order weighted-residual discretization techniques for the solution of PDEs on unstructured mesh \cite{GK_CFDbook}. In SEM simulation, the computational domain is partitioned into subdomains (elements), while the unknown variables of the PDE are approximated by high-order tensor-product polynomial expansions within each element, followed by the construction of the variational form of the PDE on the Gaussian quadrature points. Compared with low-order finite element method (FEM), SEM can achieve exponential (spectral) convergence to the exact solution by increasing the order the polynomial expansions while keeping the minimum element size fixed; compared with spectral methods, SEM can simulate physical phenomena in the complex geometry. 

Let $\Omega \subset \mathbb{R}^d$ ($d =2$ or $3$) denote the computational domain, and  $\Gamma := \partial\Omega$ denote the boundary. In Nektar++, the coupled unsteady equation \eqref{eq:problem} is solved by the high-order time-splitting scheme \cite{karniadakis1991high}. Using the test functions $\forall q \in H^1(\Omega)$ and $\forall \varphi \in H_0^1(\Omega)=\{v \in H^1(\Omega):v|_{\partial \Omega}=0\}$, where $H^1({\Omega})$ is the approximation space, the weak form of the time-splitting scheme can be formulated as follows:
\begin{equation}\label{weak_pressure}
    \int_{\Omega} \nabla p^{n+1} \cdot \nabla q = \int_{\Omega} \nabla \cdot \left( \frac{\Hat{\bm{u}}}{\Delta t}-\mathbf{N}\big( \bm{u}^{*,n+1}\big)+\bm{f}^{n+1}\right) q,
\end{equation}

\begin{equation}\label{weak_velocity}
     \frac{\gamma_0}{\nu \Delta t}\int_{\Omega} \varphi \bm{u}^{n+1}+\int_{\Omega} \nabla \varphi \cdot \nabla \bm{u}^{n+1}=
    \frac{1}{\nu} \int_{\Omega} \left(\frac{\hat{\bm{u}}}{\Delta t}-\nabla p^{n+1}\right)\varphi 
\end{equation}
where the superscript $n+1$ denotes the solution at time step $n+1$, $\bm{f}=\text{Ri}T\hat{k}$ is the body force, $\nu=\frac{1}{Re}$ is the kinematic viscosity. We set $\hat{\bm{u}}=\sum_{k=0}^{J-1}\alpha_k\bm{u}^{n-k}$ ($J$ is the order of temporal accuracy, $J=1$ or 2). $\mathbf{N}\big( \bm{u}^{*,n+1})$ is the explicit approximation of the nonlinear term, i.e., $\mathbf{N}\big( \bm{u}^{*,n+1}) \approx \mathbf{N}(\bm{u^{n+1}})=\bm{u}^{n+1}\cdot \nabla \bm{u}^{n+1}$, where $\bm{u}^{*,n+1}=\sum_{k=0}^{J-1}\beta_k\bm{u}^{n-k}$, $\alpha_k$ and $\beta_k$ are the coefficients of the stiffly-stable integrators, see \cite{GK_CFDbook}. We note that the pressure equation has to be solved with the following consistent boundary condition, 
\begin{equation}
    \frac{\partial p^{n+1}}{\partial n} = -\left[ \frac{\Hat{\bm{u}}}{\Delta t}+\sum_{k=0}^{J-1} \beta_k \mathbf{N}(\bm{u}^{n-q})+ \nu \sum_{k=0}^{J-1}\beta_k(\nabla \times \boldsymbol{\omega)}^{n-k}\right] \cdot \bm{n},
\end{equation}
where $\boldsymbol{\omega}=\nabla \times \bm{u}$ is the vorticity field, and $\bm{n}$ denotes the unit normal direction vector of the boundary face.

Since its first introduction by \cite{PATERA1984468}, SEM has undertaken significant  developments at both the fundamental and at the application {\color{cyan}levels}, which are well documented in \cite{GK_CFDbook,Deville_Fischer_Mund_2002,Hesthaven}, and several openly accessible SEM libraries have been released, e.g. Nektar++ \cite{cantwell2015nektar++}, Nek5000 \cite{nek5000-web-page}  and Libparanumal \cite{ChalmersKarakusAustinSwirydowiczWarburton2020}. Nevertheless, the use of SEM may be limited due to the following reasons: (1) the initial and boundary conditions must be noise-free, 
and (2) assimilation of sparse and/or noisy data is prohibited.

In this study, we mainly focus on two ways of integrating machine learning models, specifically PINNs, into Nektar++ to assimilate data and solve the multiphysics problem under different cases: 
\begin{enumerate}
\item Case A: replacing $\temp$ with a NN surrogate model and solve $\bm{u}$ from Eqs. \eqref{eq:problem_a} and \eqref{eq:problem_b}. 
\item Case B: replacing $\bm{u}$ with a NN surrogate model and solve $\temp$ from Eq. \eqref{eq:problem_c}. 
\item Case C: solving the entire system \eqref{eq:problem} by imposing the boundary condition using the PINNs model. 
\end{enumerate}
In all cases, NN surrogate models are obtained from training PINNs to fit the data and satisfy corresponding physics, and we denote them as $\temp_{\theta^*}$ and $\bm{u}_{\theta^*}$, respectively.
We note that by providing a NN surrogate model, which can be evaluated anywhere in the domain, and decoupling the system of \eqref{eq:problem} into Eqs. \eqref{eq:problem_a} and \eqref{eq:problem_b} and Eq. \eqref{eq:problem_c}, we are in fact solving the Navier-Stokes (NS) equation in Case A and an advection equation in Case B; in Case C we solve the coupled system \eqref{eq:problem} and impose the boundary conditions (Dirichlet, Neumann, and Robin) using PINNs to unify the solution.
We present the details of the integration as follows.

\begin{figure}
\centering
\includegraphics[trim={0cm 0cm 0cm 0cm}, clip, width=0.9\textwidth]{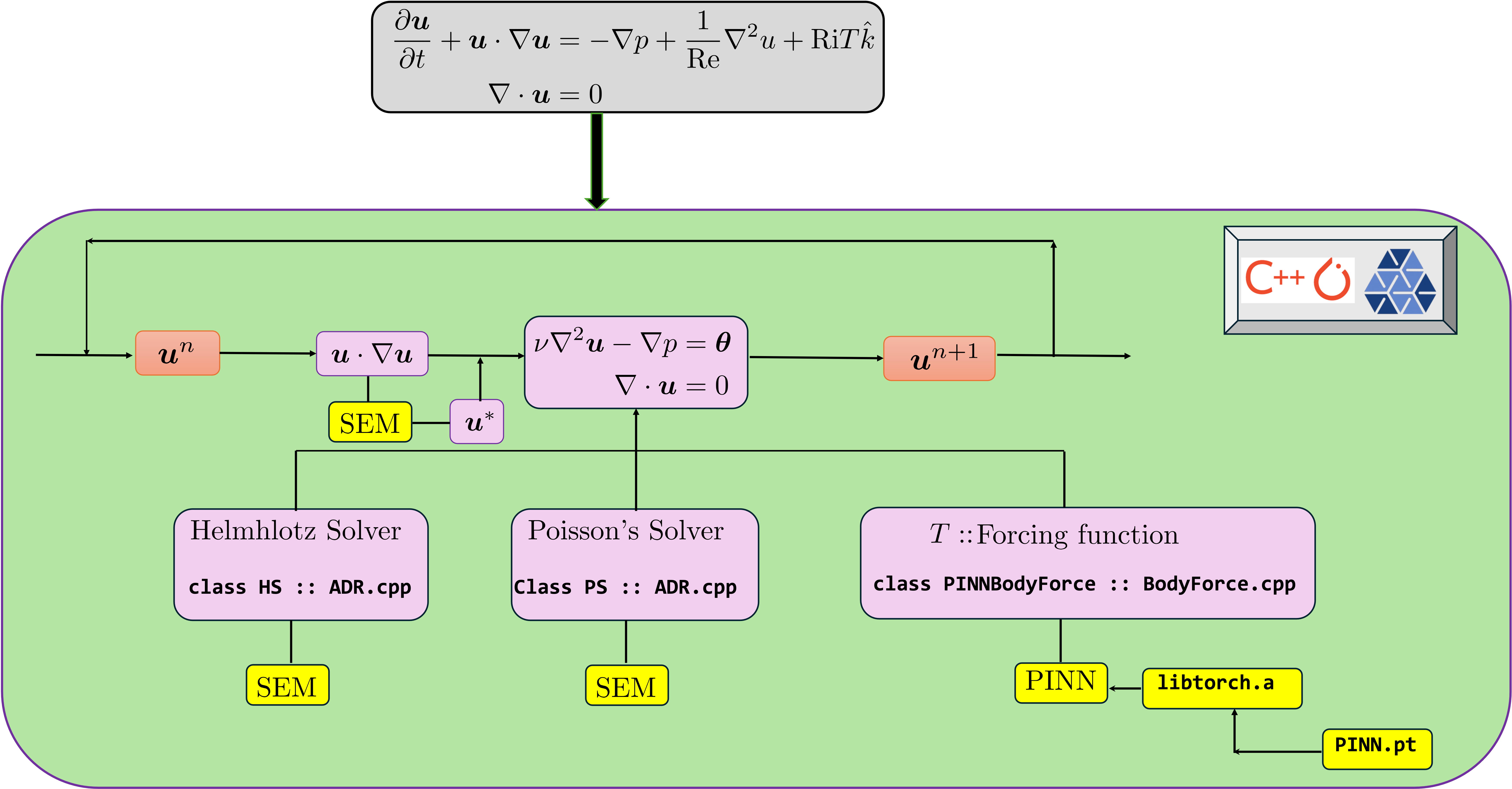}
\caption{Case A: Integration of PINNs for $\temp$ in Nektar++. The algorithm presented here to solve Eqs. \eqref{eq:problem_a} and \eqref{eq:problem_b} is motivated by the work of \cite{karniadakis1991high}.}
\label{fig:case_a}
\end{figure}
 
\subsubsection{Case A: Integration of the PINNs model for $\temp$ into Nektar++}
In this case, we solve Eqs. \eqref{eq:problem_a} and \eqref{eq:problem_b} and call the PINNs model to infer the value of the temperature $\temp$ at every time step and add it as the bodyforce. A detailed workflow for the integration performed in Case (A) is shown in Fig. \ref{fig:case_a}, which presents a typical work flow of solving the NS equation using SEM in Nektar++ . The numerical method is mainly based on an implicit-explicit approach derived by Karniadakis et al.\cite{karniadakis1991high}. For each time step, the NS solver solves a Poisson and a Helmholtz equations, which is represented by PS and HS in Fig. \ref{fig:case_a}. We add a C++ class as \texttt{PINNBodyForce.cpp}, which is a derived class of \texttt{BodyForce.cpp}.  In \texttt{BodyForce.cpp}, we call PINNs  \texttt{PINN.pt}, PyTorch traced model, and feed it with input of $x, y, t$ to compute the temperature $\temp$, and then add it as the bodyforce in Eq. \eqref{eq:problem_b}. To ease the usage of interface for seamless experience, the user has to only provide the name of the PINNs model in Session file (session.xml) of Nektar++, which includes the flow condition and mesh file (mesh.xml). An example of the session file for solving Eqs. \eqref{eq:problem_a} and \eqref{eq:problem_b} is shown in code listing \ref{code:phi_PINN} of \ref{appendix_session}. Lines 16-19 of code listing \ref{code:phi_PINN} are the changes, which the user needs to make to call the PINNs model in SEM solver. Lines 22-24 are required to update the force using the PINNs model. To run the solver, users are required to invoke the following command on the terminal:
\begin{lstlisting}[language=csh, frame=shadowbox, rulesepcolor=\color{gray}, title=Command for $\temp \rightarrow (u~ v)$]
  $ IncNavierStokesSolver mesh.xml session.xml
\end{lstlisting}

\begin{figure}
\centering
\includegraphics[trim={0cm 0cm 0cm 0cm}, clip, width=0.9\textwidth]{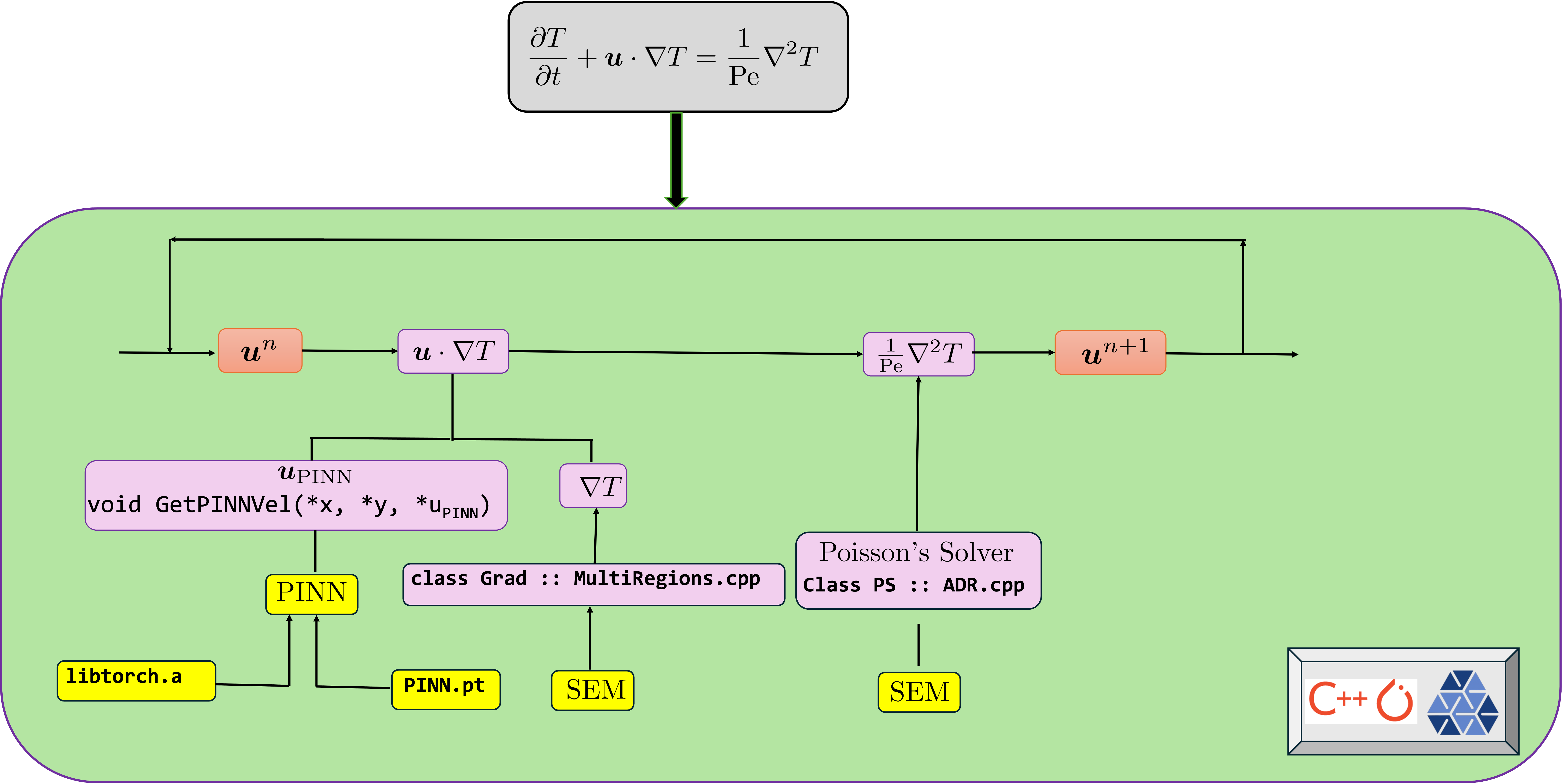}
\caption{Case (B): Integration of the PINNs  model for $\bm{u}$ in Nektar++. The integration computes $\temp$ by solving Eq. \eqref{eq:problem_c}, which is a linear advection PDE. The advection velocity $\bm{u}$ in Eq. \eqref{eq:problem_c} is computed by the pre-trained PINNs model.}
\label{fig:case_b}
\end{figure}

\subsubsection{Case B: Integration of the PINNs model for $[u, v]^T$ into Nektar++}

To integrate the PINNs model of $\bm{u} = [u, v]^T$ to compute \(\theta\), we solve only Eq. \eqref{eq:problem_c}, where the advection velocities are inferred from the PINNs model. A workflow for the integration is shown in Fig. \ref{fig:case_b}. To integrate the PINNs model in Nektar++, we modify the \texttt{ADRSolver} of Nektar++, specifically the \texttt{C++} class \texttt{UnsteadyAdvection.cpp}. The member variables \texttt{Vx} and \texttt{Vy} are initialized by inferring them from the \texttt{PINN.pt} model, which is fed with input of $x, y, t$. We add an \texttt{IF} loop to check if the parameters \(ax\) and \(ay\) are defined as ``PINN" or user-defined values. If they are defined as ``PINN", the function \texttt{FUNCTION NAME="PINNAdvectionVelocity"} is evaluated from the PyTorch \cite{paszke2019pytorch} traced model. At the usage level, the user has to make changes in session file (session.xml), which is shown in code listing \ref{code:uv_PINN} of \ref{appendix_session}. The lines 14-15 will be initialized with PINNs and lines 17-20 are to be added in the session file to infer advection velocities $(V_x=u, V_y=v)$ from the PINNs model traced by PyTorch. Finally, to run the solver a user need to type the following command on the terminal:
\begin{lstlisting}[language=csh, frame=shadowbox, rulesepcolor=\color{gray}, title=Command for $ (u~v) \rightarrow \temp $]
  $ ADRSolver mesh.xml session.xml
\end{lstlisting}

\subsubsection{Case C: Integration of the PINNs model for the boundary conditions of $[u, v]^T$ and $\temp$ into Nektar++}
In this section, we discuss the integration of the SEM solver with the PINNs model, which is trained in a small subdomain $\Omega_c$, as illustrated in Fig. \ref{fig:scenario_d_2}. PINNs provide continuous approximation of $\bm{U} = [u,~v,~\temp, ~p]^{\top}$ in $\Omega_c$. We solve Eq. \eqref{eq:problem} in the domain $\Omega_s = \Omega/ \Omega_c$ while imposing the value of $U$ from the PINNs boundary condition. To integrate the PINNs model with SEM solver, we write a parser which will infer the value of $u,~v,~\temp$ and $p$ from PINNs  at $\Omega_c$ and pass it to Nektar++ as boundary conditions on $\Omega_c$. Through  this interface we can impose the following boundary conditions on $\partial \Omega_c$
\begin{enumerate}
    \item Dirichlet boundary condition:
    \[
    \bm{U}_{\text{SEM}} |_{\partial \Omega_c} = \bm{U}_{\text{PINN}} |_{\partial \Omega_c}.
    \]

    \item Neumann boundary condition:
    \[
    \nabla \bm{U}_{\text{SEM}} \cdot \bm{n} |_{\partial \Omega_c} = \nabla\bm{U}_{\text{PINN}} \cdot \bm{n}|_{\partial \Omega_c}.
    \]

    \item Robin boundary condition:
    \[
    \nabla \bm{U}_{\text{SEM}} \cdot \bm{n} |_{\partial \Omega_c} + R \bm{U}_{\text{SEM}} |_{\partial \Omega_c} = \nabla\bm{U}_{\text{PINN}} \cdot \bm{n} |_{\partial \Omega_c} + R\bm{U}_{\text{PINN}}|_{\partial \Omega_c}. 
    \]
\end{enumerate}

In the Robin boundary condition when the coefficient $R$ is zero we recover the Neumann boundary condition. However, If $R$ approaches  infinity, the Robin boundary condition approaches the Dirichlet boundary condition.
The more general boundary condition is the Robin one but the coefficient $R$ has to be optimized for accurate results.
The usage of integrated framework can be invoked by the following command
\begin{lstlisting}[language=csh, frame=shadowbox, rulesepcolor=\color{gray}, title=Command for Integration of PINN model for the boundary conditions]
  $ IncNavierStokesSolver mesh.xml session.xml
\end{lstlisting}
An example of the session file for this integration is shown in code listing \ref{code:cutout_PINN} of \ref{appendix_session}.

{\color{blue}
\subsection{NeuroSEM versus existing data assimilation approach}

NeuroSEM is fundamentally different from conventional data assimilation methods such as 3D-VAR and 4D-VAR \cite{law2015data, fletcher2022data, evensen2022data, gelfand2010handbook, carrassi2018data, reichle2008data, bouttier2002data, lahoz2010data, asch2016data}. For instance, 3D-VAR seeks to find an optimal model state by minimizing the difference between model predictions and available observations at a specific time. It adjusts the model's state variables in three dimensions at a single time step, without explicitly considering the system's time evolution, assuming that observations and model states are valid simultaneously. In contrast, NeuroSEM seamlessly incorporates both temporal and spatial dimensions simultaneously. This enables us to handle unsteady flow around a cylinder as well as PIV data for any flow.
However, from an implementation standpoint, 3D-VAR minimizes a cost function that balances the difference between the model and observations (weighted by their uncertainties) with a background model forecast. As a result, having a background model is a core requirement, which is typically unavailable in most cases except for problems related to climate modeling.
However, while 4D-VAR can be applied to time-dependent systems, it relies on the adjoint of the numerical model, which calculates the gradient of the cost function with respect to the initial conditions. Developing and maintaining an accurate adjoint model is complex, error-prone, and computationally demanding. Moreover, adjoint models are challenging to implement for nonlinear systems or models with physical parameterizations (e.g., turbulence, convection). In contrast, NeuroSEM can effectively handle nonlinear systems seamlessly, as PDEs are encoded in PINNs using an efficient and precise automatic differentiation method \cite{baydin2018automatic, shukla2023randomized}. Finally, 4D-VAR is most suitable for weakly nonlinear systems, as strong nonlinearities in the model or observation operator can lead to difficulties in the convergence of the minimization process.
}

\section{Computational experiments}\label{sec:3}

\begin{figure}[h!]
    \centering
    \subfigure[The steady-state flow example]{
    \includegraphics[scale=.15]{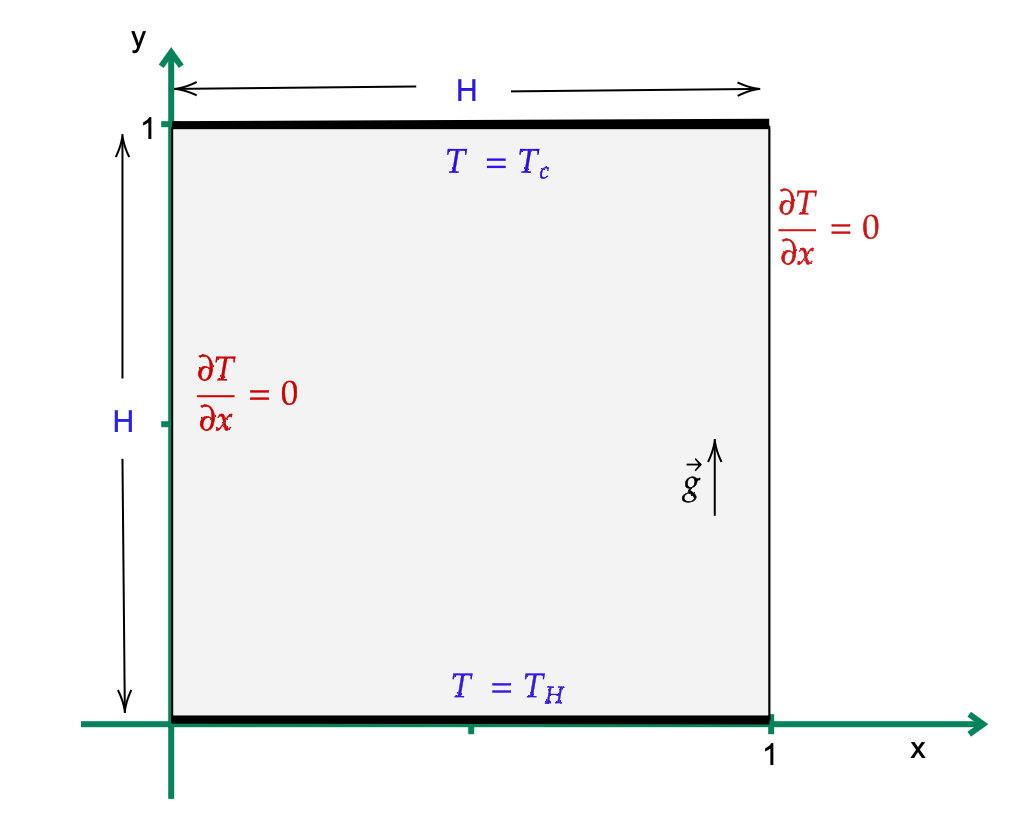}
    }
    \subfigure[The unsteady flow example]{
    \includegraphics[scale=.2]{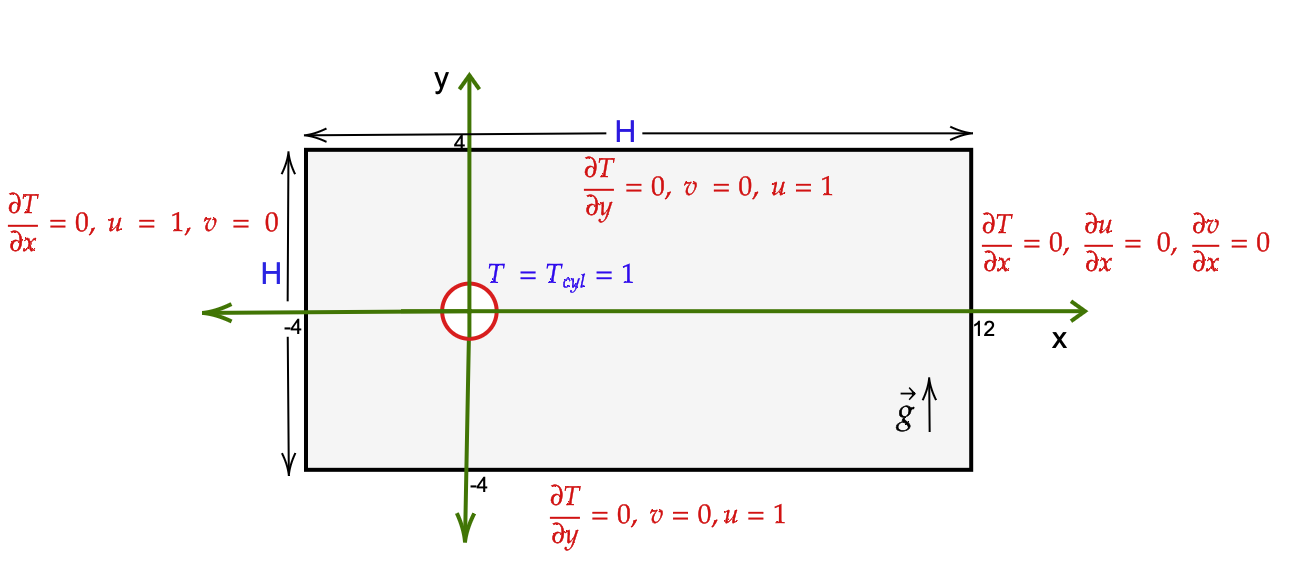}
    }
    \caption{Problem setups for the Rayleigh-B\'enard convection \eqref{eq:problem} with buoyancy driven thermal convection.}
    \label{fig:domain}
\end{figure}

In this work, we test the proposed NeuroSEM method, which integrates PINNs into Nektar++ for solving the multiphysics problem \eqref{eq:problem}, on the following two examples: (1) steady-state flows on a square domain in Sec. \ref{sec:3_1}, and (2) an unsteady flow past a cylinder in Sec. \ref{sec:3_2}. The problem setups including the computational domains are displayed in Fig. \ref{fig:domain}. The details regarding the hyperparameter and the training of PINNs can be found in \ref{sec:appendix_pinn}. The code for reproducing the results of PINNs will be made publicly available at \url{https://github.com/ZongrenZou/NeuroSEM} upon the acceptance of the paper. We remark that in some cases we will refer to SEM as the ``reference method",
meaning that the SEM providing the reference value is the one applied to solve the entire coupled multiphysics problem, rather than the SEM component in the NeuroSEM method.

\subsection{A steady-state flow example}\label{sec:3_1}

For the steady-state example, we solve the steady-state of Eq. \eqref{eq:problem} on a square domain, defined as $\Omega = [0, 1]^2$ (shown in Fig. \ref{fig:domain}(a)). Here the unit normal vector and the non dimensional temperature are defined as $\bm{k} = [0, 1]^\top$ and
\begin{equation}
    \temp = \frac{\tilde{T} - T_r}{T_H - T_C},
\end{equation}
respectively, where $\tilde{T}, T_H, T_C, T_r = \frac{T_H + T_c}{2}$ refer to the temperature, the hot temperature along the bottom wall ($y=0$), cold temperature along the top walls ($y=1$), and reference temperature, respectively.
The Rayleigh and Prandtl numbers are given as $\text{Ra} = \frac{\alpha g \Delta\temp H^3}{\kappa \nu}$ and $\text{Pr} = \frac{\nu}{\kappa}$, respectively, where $H=1$ and $\alpha, g, \Delta\temp, \nu, \kappa$ refer to the thermal expansion coefficient, acceleration due to gravity, the non-dimensional temperature difference between the bottom and top wall ($\Delta\temp = \temp_{y=0} - \temp_{y=1}$), kinematic viscosity, and thermal diffusivity of the fluid, respectively. The relation between $\text{Ra}$ and $\text{Pr}$ is defined as:
\begin{equation}
    \text{Ra} = \text{Gr}\text{Pr},
\end{equation}
where $\text{Gr}$ is Grashof number and defined as $\text{Gr} = \text{Ri}{\text{Re}^2}$. 
{\color{cyan}In our case, the P\'eclet number $\text{Pe} \gg 1$, indicating that there is a strong coupling wherein buoyancy forces are significant, leading to rigorous convection.} For the Rayleigh-B\'{e}nard problem setup, we will compare the Nusselt number, $Nu=\frac{\partial \temp}{\partial y}|_{y=0} / \frac{\Delta \temp}{H}$, which measures the ratio between convective and conductive heat transfer, across different scenarios.

In this example, we specifically consider the following four scenarios for different data availability and targets:
\begin{enumerate}
    \item \textbf{Scenario A}: Given some data of $\bm{u}=(u, v)$ at scattered points across the domain $\Omega$, we aim to obtain continuous fields $\bm{u}(x, y)$ and $\temp(x, y)$ via SEM and PINNs, respectively.
    \item \textbf{Scenario B}: 
    Given some data of $\temp$ at scattered points across the domain $\Omega$, we aim to obtain continuous fields $\bm{u}(x, y)$ and $\temp(x, y)$ via PINNs and SEM, respectively.
    \item \textbf{Scenario C}:
    Given some data of $\bm{u}=(u, v)$ and $\temp$ at scattered points across the domain $\Omega$ but we have unknown thermal boundary conditions on $\partial\Omega$, we aim to obtain continuous fields $\bm{u}(x, y)$ and $\temp(x, y)$ via SEM and PINNs, respectively,  as well as recover the missing thermal boundary conditions.
    \item \textbf{Scenario D}:
    Given some data of $\bm{u}=(u, v)$ and $\temp$ at scattered points in a small subdomain, we aim to obtain continuous fields $\bm{u}(x, y)$ and $\temp(x, y)$ across the entire domain $\Omega$ via PINNs in the subdomain and SEM in the remaining domain.
\end{enumerate}
We note that the first two scenarios (\textbf{Scenario A} and \textbf{Scenario B}) are only for the verification of the NeuroSEM method.
One can use SEM to solve the coupled system directly because the boundary condition of the multiphysics problem is fully specified, while in the rest two scenarios (\textbf{Scenario C} and \textbf{Scenario D}), SEM cannot be directly applied because of the missing boundary conditions.

\begin{figure}
\centering
\includegraphics[width=\textwidth]{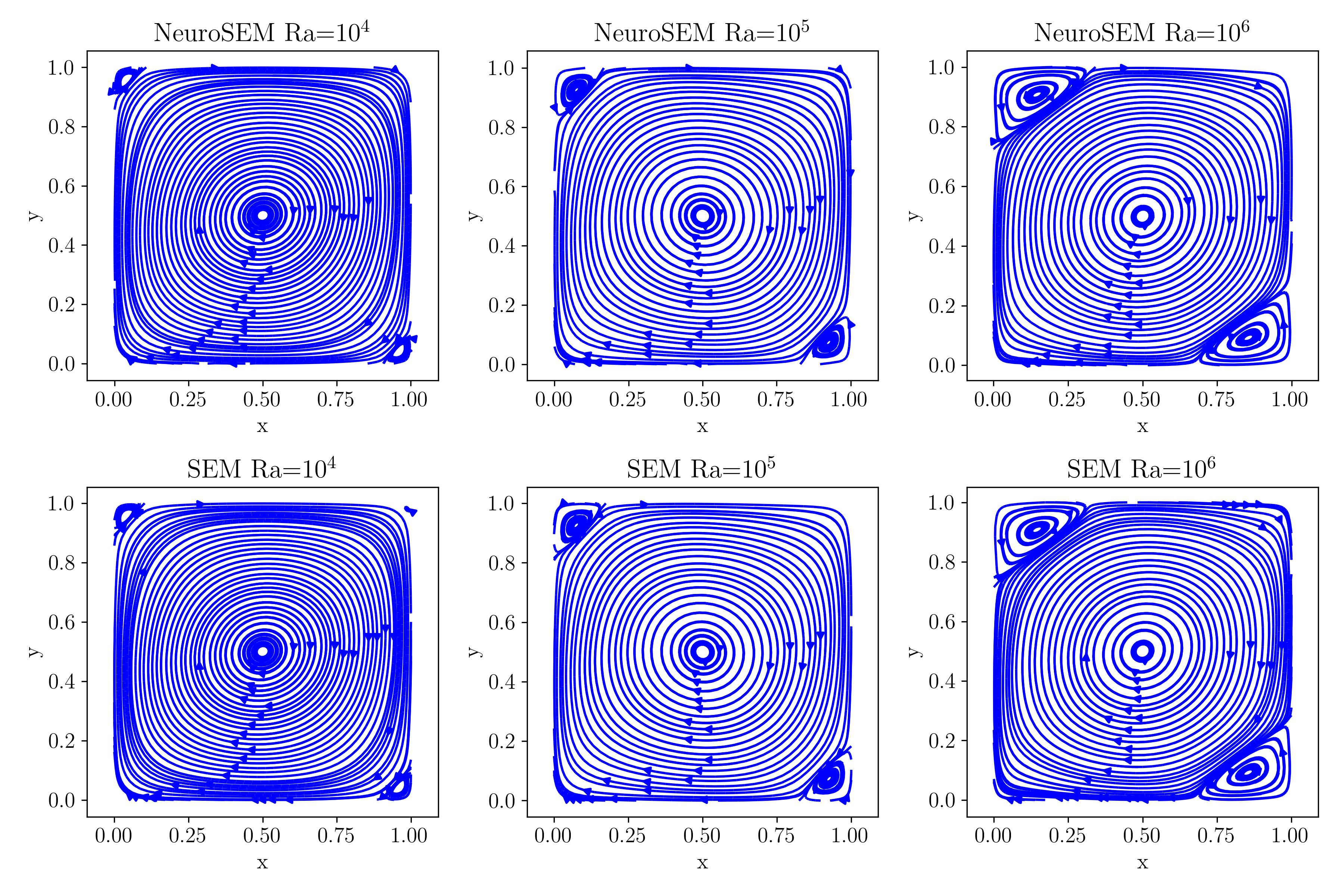}
\caption{\textbf{Scenario A}: Streamline plots obtained from NeuroSEM (top row) and SEM (bottom row). The left, center and right subfigures show the streamline plots for $\text{Ra}=10^4$, $10^5$ and $10^6$, respectively. 
The errors of $\bm{u} = [u, v]^T$ are presented in Table \ref{tab:scenario_a}.}
\label{fig:scenario_a}
\end{figure}

\begin{table}[ht!]
    \footnotesize
    \centering
    \begin{tabular}{c|c|c|c}
    \hline\hline
     & $\text{Ra}=10^4$ & $\text{Ra}=10^5$ & $\text{Ra}=10^6$ \\
    \hline
    Relative $L_2$-error of $u$ & $0.090\%$ & $0.181\%$ & $0.336 \%$  \\
    \hline
    Relative $L_2$-error of $v$ & $0.091\%$ & $0.176\%$ & $0.336\%$ \\
    \hline\hline
    \end{tabular}
    \caption{\textbf{Scenario A}: The relative $L_2$-error of $\bm{u} = [u, v]^T$ from the NeuroSEM method for different Rayleigh numbers ($\text{Ra}$). }
    \label{tab:scenario_a}
\end{table}

In \textbf{Scenario A}, we assume that scattered data of $\bm{u}$ are available across $\Omega$, and we assimilate these data by first solving $\temp$ from Eq. \eqref{eq:problem_c} using the PINNs method to obtain a NN surrogate for $\temp$.
Specifically, we have $N_{\bm{u}}=10,000$ data of $\bm{u}$, $\{x_i, y_i, \bm{u}_i\}_{i=1}^{N_{\bm{u}}}$, randomly sampled from the domain $\Omega$, and construct the PINNs loss function $\mathcal{L}(\theta)$, as follows:
\begin{equation}\label{eq:loss_scenario_a}
    \begin{aligned}
        \mathcal{L}(\theta) = & \frac{w_{\bm{u}}}{N_{\bm{u}}}\sum_{i=1}^{N_{\bm{u}}}|\bm{u}_i\cdot \nabla \temp_\theta(x_i, y_i) - \frac{1}{\text{Pe}}\nabla^2\temp_\theta(x_i, y_i)|^2 + w_b\mathcal{L}_{b, \temp_\theta}(\theta)
    \end{aligned}
\end{equation}
where $\mathcal{L}_{b, \temp_\theta}(\theta)$ is the loss function for the boundary conditions of $\temp$, and $w_{\bm{u}}$ and $w_{b}$ are weighting coefficients for different terms in the loss function.  
We note that in this scenario, the NN surrogate takes as input $x, y$ and outputs $\temp$, and we are in fact solving an advection-diffusion system using the PINNs method. 
The trained NN model, denoted as $\temp_{\theta^*}$ where $\theta^*$ denotes the minimizer of the loss function \eqref{eq:loss_scenario_a} obtained from solving the optimization problem with certain optimizers, e.g. Adam \cite{kingma2014adam}, is then fed as a forcing function to the SEM solver to solve the NS equations of $\bm{u}$, i.e. Eqs. \eqref{eq:problem_a} and \eqref{eq:problem_b}, with the boundary conditions of $\bm{u}$. The workflow of \textbf{Scenario A} is illustrated in Fig. \ref{fig:case_a}.
We present the results of the NeuroSEM method in Fig. \ref{fig:scenario_a}, from which we observe that the streamline plots obtained from NeuroSEM for different Ra are shown to agree with the ones obtained from the reference method, which solves the coupled problem \eqref{eq:problem} using SEM.
Furthermore, our approach is able to recover {\color{cyan}the correct} vortices around the corner, which are often difficult to be reconstructed from PINNs as the lone framework, especially when $\text{Ra}$ is large.
The errors of $\bm{u}=[u, v]^T$ from our approach for different Rayleigh numbers are also presented in Table \ref{tab:scenario_a}. We can see that the errors are small and increase as the Rayleigh number grows.
In particular for this scenario, we have conducted a comprehensive study over variants of PINNs, e.g. separable PINNs \cite{cho2023separable} and self-adaptive PINNs \cite{mcclenny2020self}, for the purpose of accuracy and efficiency (see \ref{accl_pinn}).

\begin{figure}[h!]
    \centering
    \subfigure[$\text{Ra}=10^4$]{
    \includegraphics[scale=.28]{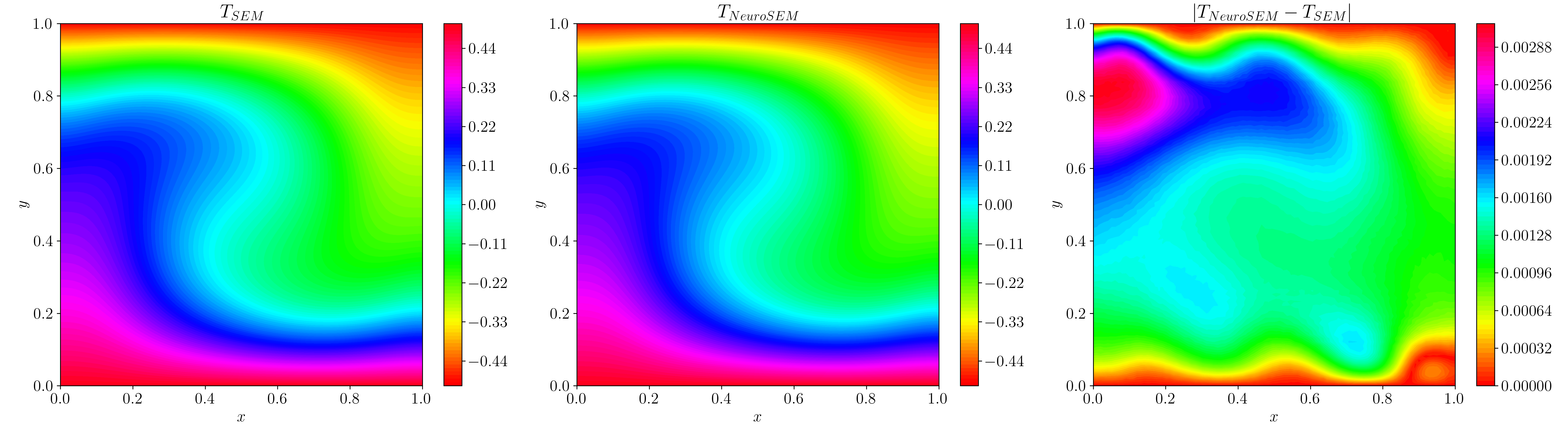}
    }
    \subfigure[$\text{Ra}=10^5$]{
    \includegraphics[scale=.28]{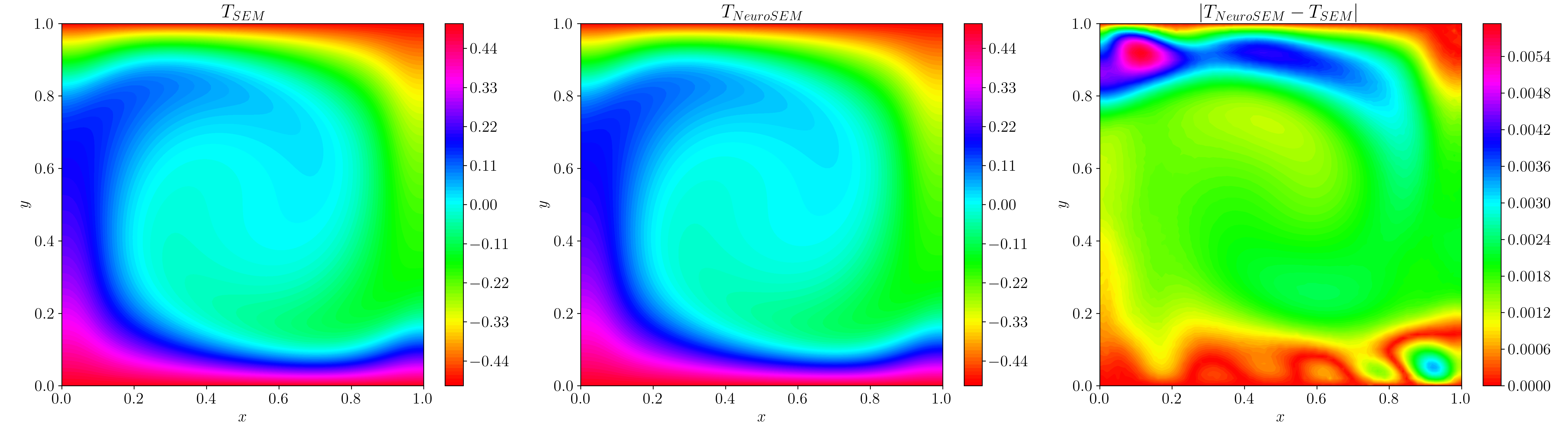}
    }
    \subfigure[$\text{Ra}=10^6$]{
    \includegraphics[scale=.28]{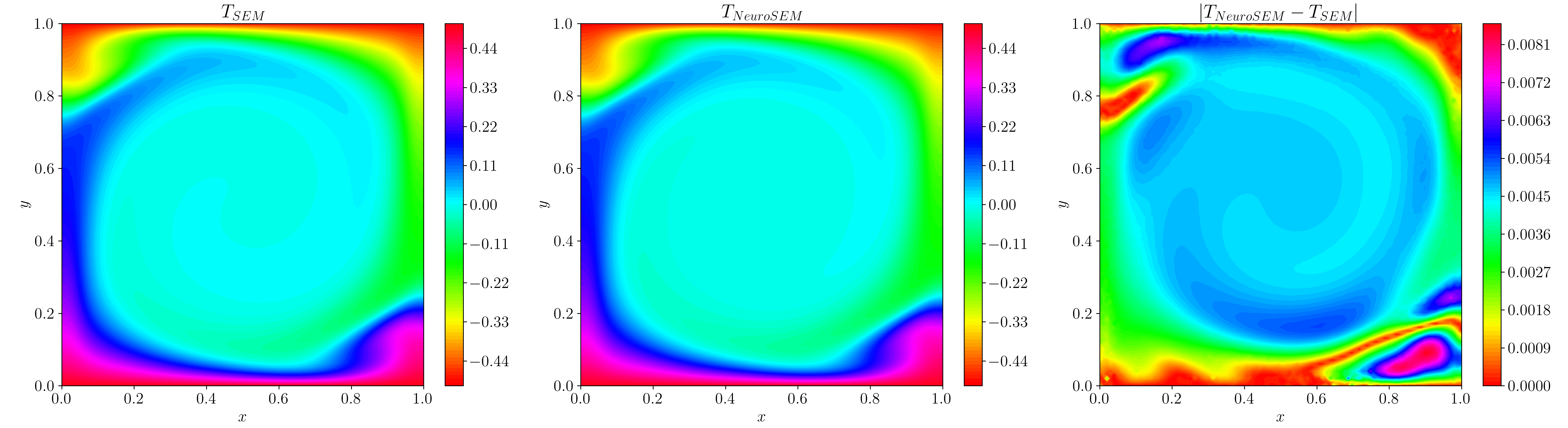}
    }
    \caption{\textbf{Scenario B}: Reconstructed $\temp$ from NeuroSEM and SEM. The top, middle, and bottom rows show $\temp$ plots for $\text{Ra}=10^4$, $\text{Ra}=10^5$ and $\text{Ra}=10^6$, respectively. To validate the results, we also present plots of $\temp$ obtained from SEM by solving \eqref{eq:problem}. The errors of $T$ are presented in Table \ref{tab:scenario_b}.}
    \label{fig:scenario_b_1}
\end{figure}

\begin{figure}
    \centering
    \includegraphics[width=\textwidth]{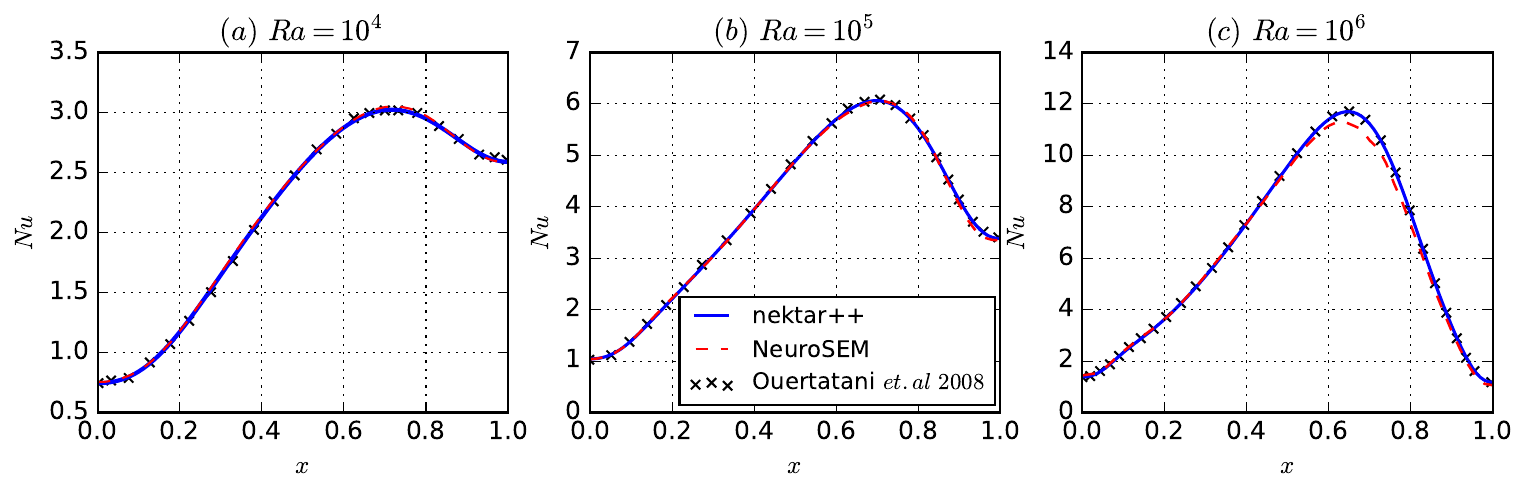}
    \caption{\textbf{Scenario B}: Nusselt number ($Nu = \frac{\partial \temp}{\partial y}|_{y=0}/\frac{\Delta \temp}{H}$) comparison between Nektar++, NeuroSEM and Ouetatani et.al \cite{ouertatani2008numerical} for (a) $\text{Ra}=10^4$, (b) $\text{Ra} = 10^5$, (c) $\text{Ra} = 10^6$ from left to right.} 
    \label{fig:scenario_b_2}
\end{figure}

\begin{table}[ht!]
    \footnotesize
    \centering
    \begin{tabular}{c|c|c|c}
    \hline\hline
     & $\text{Ra}=10^4$ & $\text{Ra}=10^5$ & $\text{Ra}=10^6$ \\
    \hline
    Relative $L_2$-error of $T$ & $0.57\%$ & $1.02\%$ & $2.43 \%$  \\
    \hline\hline
    \end{tabular}
    \caption{\textbf{Scenario B}: The relative $L_2$-error of $T$ from the NeuroSEM method for different Rayleigh numbers ($\text{Ra}$). }
    \label{tab:scenario_b}
\end{table}

In \textbf{Scenario B}, we assume that we have scattered data of $\temp$ across $\Omega$, and we assimilate these data by first solving $\bm{u}$ from Eqs. \eqref{eq:problem_b} and \eqref{eq:problem_b} using the PINNs method to obtain a NN surrogate for $\bm{u}$. Specifically, we have $N_\temp=5,000$ data of $\temp$, $\{x_i, y_i, \temp_i\}_{i=1}^{N_\temp}$, randomly sampled from the domain, and construct the PINNs loss function $\mathcal{L}(\theta)$ as follows:
\begin{equation}
    \begin{aligned}
        \mathcal{L}(\theta) = &\frac{w_\temp}{N_\temp}\sum_{i=1}^{N_\temp} ||(\bm{u}_\theta\cdot \nabla\bm{u}_\theta)(x_i, y_i) + \nabla p_\theta(x_i, y_i) - \frac{1}{\text{Re}}(\nabla ^2 \bm{u}_\theta)(x_i, y_i) - \text{Ri}\temp_i \hat{\bm{k}}||_2^2 + \\
        &\frac{w_f}{N_f}\sum_{i=1}^{N_f}|(\nabla\cdot \bm{u}_\theta)(x_i^f, y_i^f)|^2 + \mathcal{L}_{b, \bm{u}_\theta}(\theta),
    \end{aligned}
\end{equation}
where $x_i^f, y_i^f, i=1,...,N_f$ are residual points for Eq. \eqref{eq:problem_a} and $\mathcal{L}_{b, \bm{u}}(\theta)$ is the regular PINNs loss function for the boundary conditions of $\bm{u}$. Here, we set the residual points for Eq. \eqref{eq:problem_a} to be the same as the points where we have data of $\temp$. We note that in this scenario the NN surrogate takes as input $x, y$ and outputs $\bm{u}, p$ where $p$ denotes the pressure.
The trained NN surrogate for $\bm{u}$, denoted as $\bm{u}_{\theta^*}$, is then fed as a continuous and differentiable function to evaluate the advection velocity ($\bm{u}$) to the SEM solver to solve the advection-diffusion system of $\temp$, i.e. Eq. \eqref{eq:problem_c}, with the boundary conditions of $\temp$. The workflow of \textbf{Scenario B} is illustrated in Fig. \ref{fig:case_b}.
Similarly as in the previous scenario, we test our approach for $\text{Ra} = 10^4, 10^5, 10^6$ and compare the reconstructed temperature $\temp$ from NeuroSEM with the reference obtained from solving the coupled problem \eqref{eq:problem} using SEM. 
The reconstructions of $\temp$ and their accuracy are presented in Fig. \ref{fig:scenario_b_1} and Table \ref{tab:scenario_b}, showing the high accuracy of the NeuroSEM method in addressing different Rayleigh numbers. We also observe that the error of $\temp$ increases with the Rayleigh number.
We further compare the Nusselt number, $Nu=\frac{\partial \theta}{\partial y}|_{y=0} / \frac{\Delta\temp}{H}$, for this scenario, and the result is shown in Fig. \ref{fig:scenario_b_2}. In Fig. \ref{fig:scenario_b_2}, we plot the Nusselt numbers for \(\text{Ra} = 10^4,~10^5,~10^6\) obtained from Nektar++ and NeuroSEM, and compare them against the results of \cite{ouertatani2008numerical}. The results from NeuroSEM show excellent agreement with both SEM and \cite{ouertatani2008numerical}.

\begin{figure}
\centering
\includegraphics[width=\textwidth]{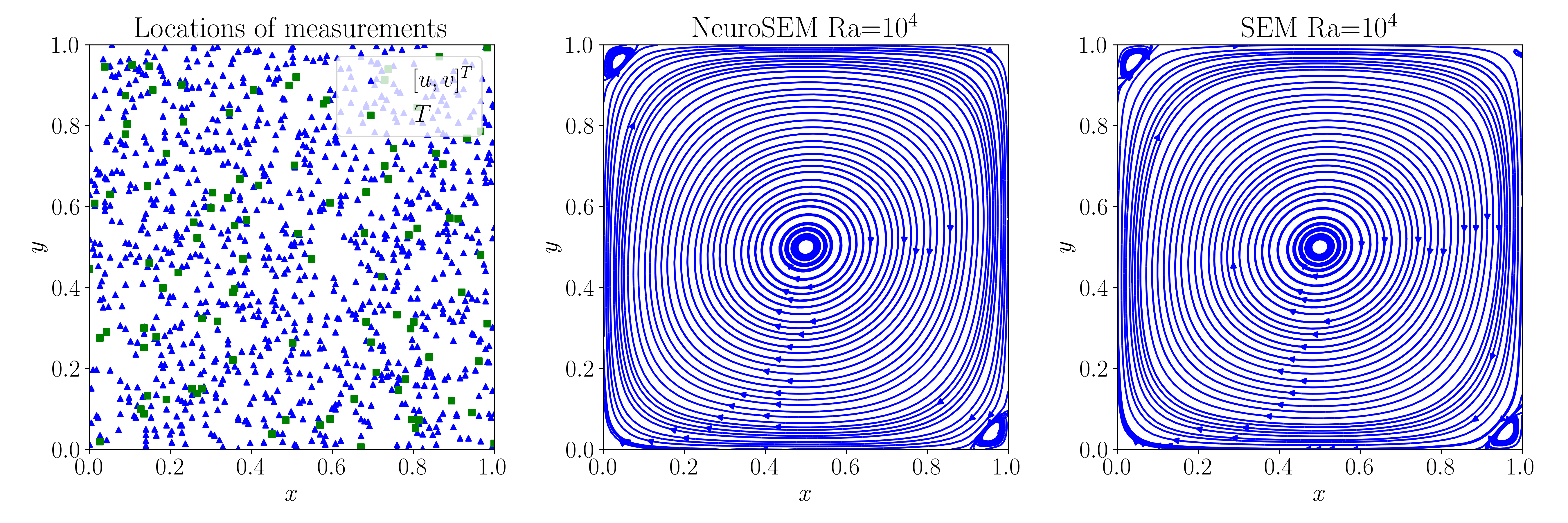}
\caption{\textbf{Scenario C}: Streamline plots for $\text{Ra}=10^4$ obtained from NeuroSEM and SEM. The left figure shows the location of noisy data points for $(u, v)$ (blue), taken as residual points, and $\temp$ (green), which completely lie inside the domain and no data on the boundary are considered. The data points are corrupted by 0.01 additive Gaussian noise. The figure in the center shows the streamline plot obtained from NeuroSEM. To validate these results we present corresponding streamline plots obtained from SEM by solving Eq. \eqref{eq:problem} using SEM with reference results shown in the right figure. The relative $L_2$-errors for $u$ and $v$ are $0.87\%$ and $0.93\%$, respectively.}
\label{fig:scenario_c}
\end{figure}

In \textbf{Scenario C}, we consider the case where we have noisy data of $\bm{u}$ and $\temp$ at scattered points across the domain but the boundary conditions of $\temp$ are not fully specified. Under this circumstance, we are unable to employ the SEM solver because of the missing boundary condition.
Without loss of generality, we assume that the Dirichlet boundary conditions of $\temp$ are missing while the Neumann boundary conditions are known. 
Specifically, we have $N_{\bm{u}}=1,000$ measurements of $\bm{u}$, $\{x^u_i, y^u_i, \bm{u}_i\}_{i=1}^{N_{\bm{u}}}$, and $N_{\temp}=100$ measurements of $\temp$, $\{x_i^\temp, y_i^\temp, \temp_i\}_{i=1}^{N_\temp}$, both of which are randomly sampled from the domain $\Omega$ (the locations of these measurements are displayed in Fig. \ref{fig:scenario_c}) and corrupted by additive Gaussian noises of mean zero and scale $0.01$. {\color{blue}We observe that the average absolute values of the uncorrupted data are $0.0936$, $0.0899$, and $0.2062$ for $u$, $v$, and $T$, respectively. Therefore, the relative magnitudes of the additive noise are approximately $10\%$, $10\%$, and $5\%$ for each variable, respectively.}
The PINNs loss function $\mathcal{L}(\theta)$ is constructed as follows:
\begin{equation}\label{eq:loss_scenario_c}
    \begin{aligned}
        \mathcal{L}(\theta) = & \frac{w_{\bm{u}}}{N_{\bm{u}}}\sum_{i=1}^{N_{\bm{u}}}|\bm{u}_i\cdot \nabla \temp_\theta(x^u_i, y^u_i) - \frac{1}{\text{Pe}}\nabla^2\temp_\theta(x^u_i, y^u_i)|^2 + \\
        & \frac{w_\temp}{N_\temp} \sum_{i=1}^{N_\temp} |\temp_\theta(x^\temp_i, y^\temp_i) - \temp_i|^2 + w_b\mathcal{L}_{Neumann, \temp_\theta}(\theta),
    \end{aligned}
\end{equation}
where $\mathcal{L}_{Neumann, \temp_\theta}(\theta)$ denotes the loss function for the Neumann boundary conditions of $\temp$. The trained NN surrogate $\temp_{\theta^*}$ is then fed to the SEM solver as a forcing function to solve the NS equations of $\bm{u}$, i.e. Eqs. \eqref{eq:problem_a} and \eqref{eq:problem_b}.
The workflow of \textbf{Scenario C} for the integration is in fact the same as the one of \textbf{Scenario A}. 
Results for $\text{Ra} = 10^4$ are displayed in Fig. \ref{fig:scenario_c}. We can see that given noisy data, our approach is able to reconstruct the velocity field with relative small errors: relative $L_2$-errors are $0.87\%$ and $0.93\%$ for $u$ and $v$, respectively.
{\color{blue}To test the robust performance of NeuroSEM in handling noisy data, we further consider four different noise levels of the additive Gaussian noise in obtaining data of $\bm{u}$ and $T$: (1) $0.01$ for $\bm{u}$ and $0.025$ for $T$, (2) $0.025$ for $\bm{u}$ and $0.01$ for $T$, (3) $0.01$ for $\bm{u}$ and $0.05$ for $T$, and (4) $0.05$ for $\bm{u}$ and $0.01$ for $T$. The results of NeuroSEM for these four cases are presented in Table \ref{tab:scenario_c}.
We note that, in this scenario, increasing the noise scale for the $T$ data leads to a more significant performance degradation in the accuracy of NeuroSEM compared to the data of $\bm{u}$.}

\begin{table}[ht!]
    \footnotesize
    \centering
    \begin{tabular}{c|c|c|c|c}
    \hline\hline
     & Noise level (1) & Noise level (2) & Noise level (3) & Noise level (4) \\
    \hline
    Relative $L_2$-error of $u$ & $2.95\%$ & $0.63\%$ & $6.06\%$ & $1.31\%$ \\
    \hline
    Relative $L_2$-error of $v$ & $3.14\%$ & $0.68\%$ & $6.44\%$  & $1.92\%$  \\
    \hline\hline
    \end{tabular}
    \caption{{\color{blue}\textbf{Scenario C}: The relative $L_2$-error of $\bm{u}$ from the NeuroSEM method for four noise levels: (1) $0.01$ for $\bm{u}$ and $0.025$ for $T$, (2) $0.025$ for $\bm{u}$ and $0.01$ for $T$, (3) $0.01$ for $\bm{u}$ and $0.05$ for $T$, and (4) $0.05$ for $\bm{u}$ and $0.01$ for $T$. }}
    \label{tab:scenario_c}
\end{table}

\begin{figure}[ht!]
\centering
\includegraphics[width=1\textwidth]{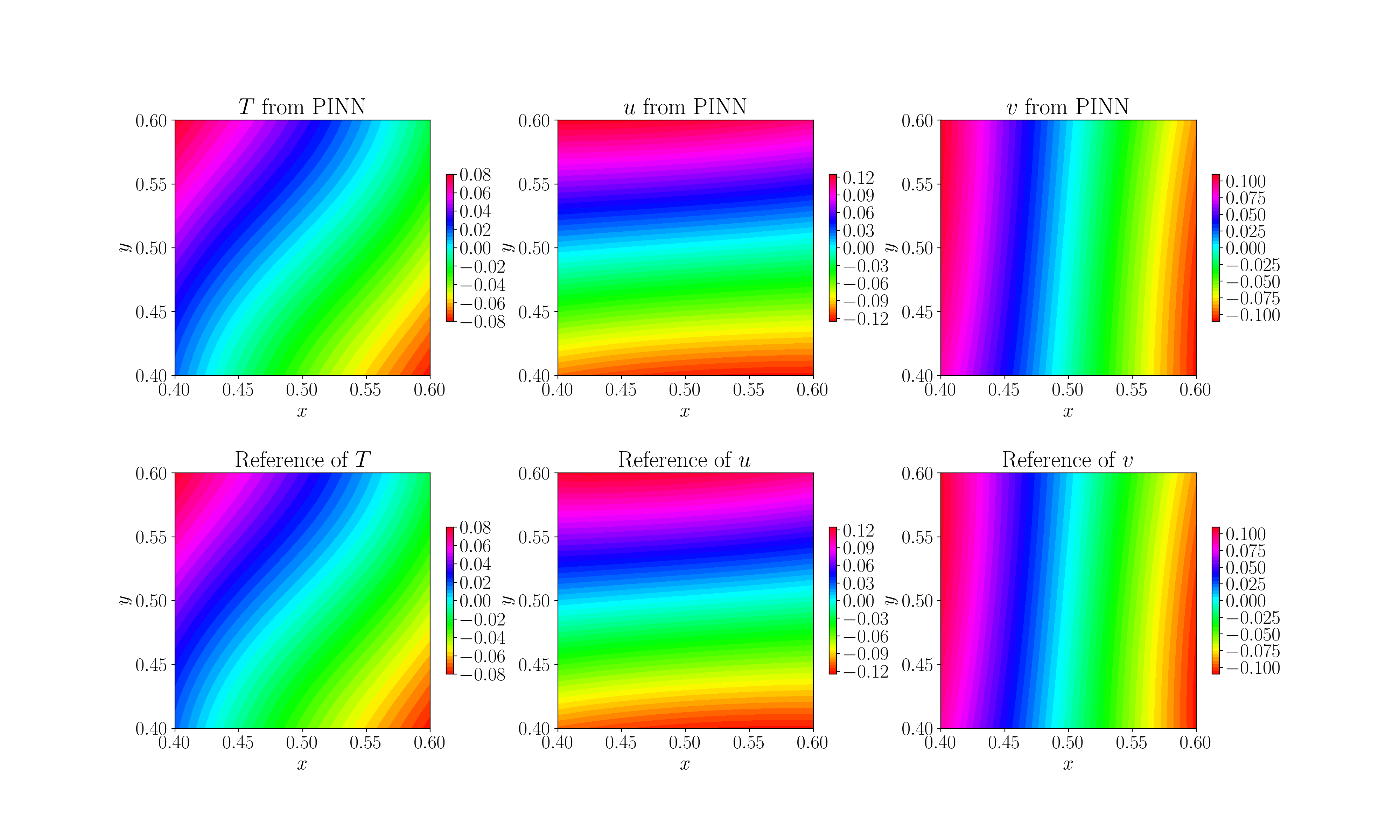}
\caption{\textbf{Scenario D}: Results of $\bm{u}=[u, v]^T$ and $\temp$ from employing PINNs on a cut-out $\Omega_c = [0.4, 0.6]^2$ with data of $\bm{u}$ and $\temp$ for $\text{Ra}=10^4$ (top row) and the references (bottom row) obtained from SEM.}
\label{fig:scenario_d_1}
\end{figure}

\begin{figure}[ht!]
\centering
\includegraphics[width=0.7\textwidth]{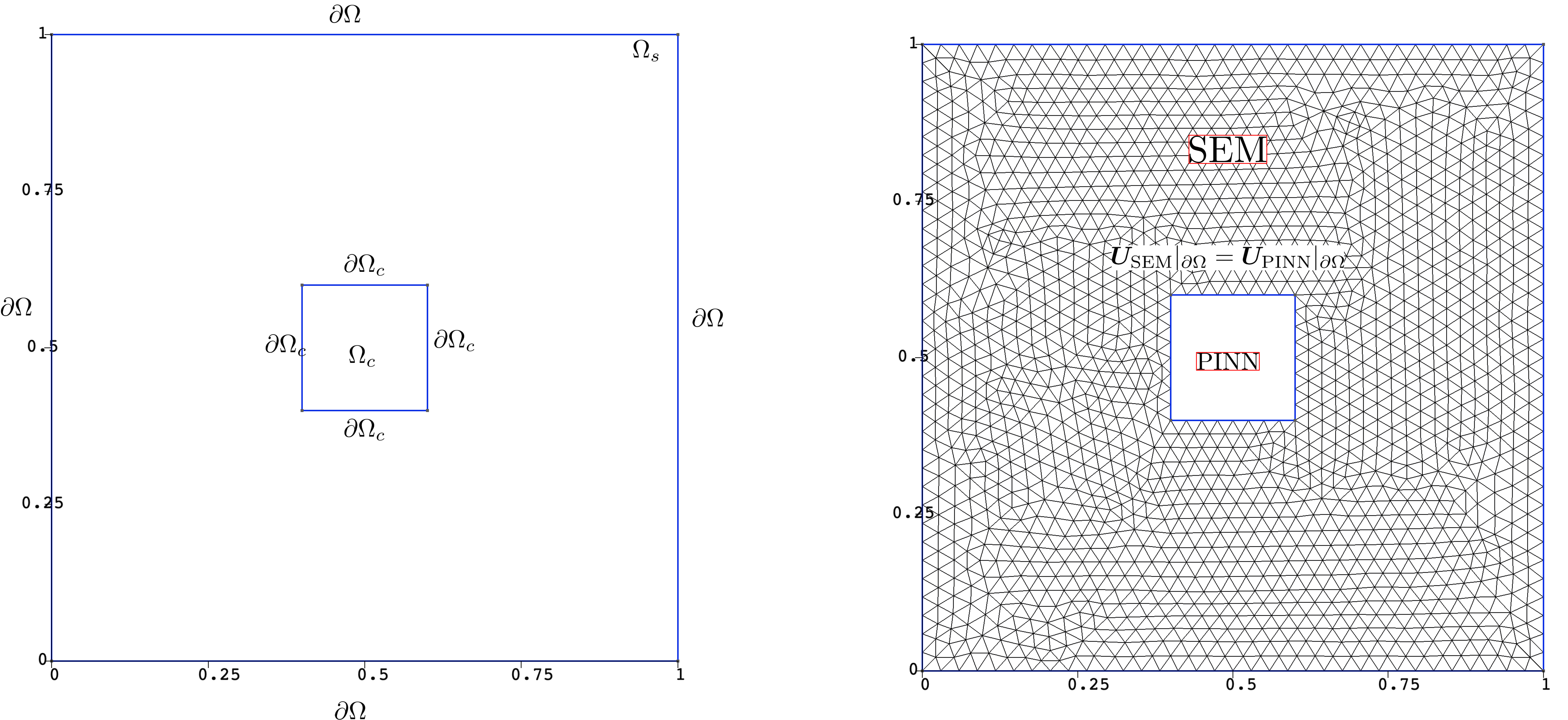}
\caption{\textbf{Scenario D}: The computational domain and boundary conditions for NeuroSEM utilize a surrogate model to evaluate the flow field in \(\Omega_c\). The left subfigure illustrates the domain \((\Omega_s = \Omega/\Omega_c)\) where NeuroSEM is executed, with boundary conditions prescribed on \(\partial \Omega\) and \(\partial \Omega_c\) (from the PINNs model). The right subfigure depicts the discretization of the fluid domain with triangles and the Dirichlet boundary condition on \(\Omega_c\).}
\label{fig:scenario_d_2}
\end{figure}

\begin{figure}[ht!]
\centering
\includegraphics[trim={5cm 0cm 5cm 0cm}, clip, width=1\textwidth]{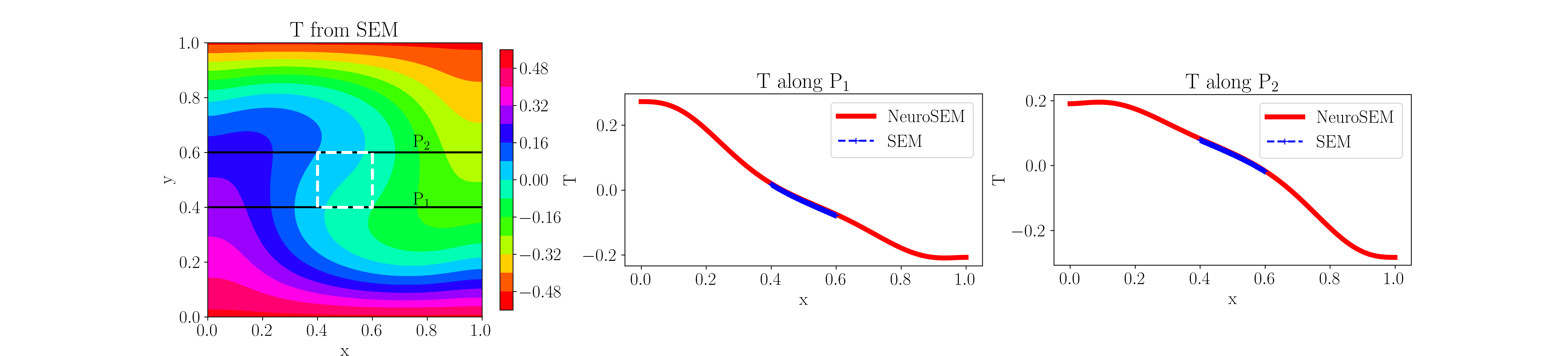}
\caption{\textbf{Scenario D}: A comparison between the solutions obtained from SEM and NeuroSEM for \(\text{Ra}=10^4\). The left subfigure shows the temperature \(\temp\) obtained from the SEM solver, with a white dashed box indicating the position of the cut-out. The middle and rightmost subfigures display the temperature profiles obtained from NeuroSEM, overlaid with the temperature (blue dots) obtained from SEM in the cut-out region, along profiles P1 and P2, respectively. }
\label{fig:scenario_d_3}
\end{figure}

\begin{figure}
    \centering
    \includegraphics[width=\textwidth]{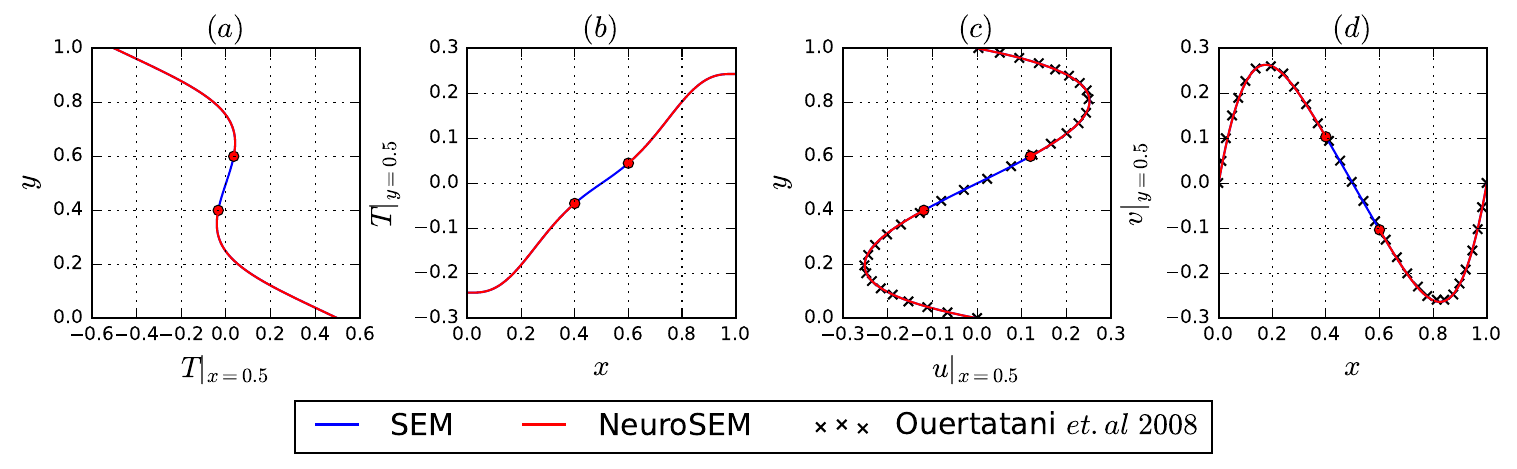}
    \caption{\textbf{Scenario D}: A comparison of results obtained from NeuroSEM, SEM and \cite{ouertatani2008numerical} for (a) $\temp(x=0.5, y)$, (b) $\temp(x, y=0.5)$, (c) $u(x=0.5, y)$, (d) $v(x, y=0.5)|$}
    \label{fig:scenario_d_4}
\end{figure}

In \textbf{Scenario D}, we assume that we have scattered data of $\bm{u}$ and $\temp$ inside a cut-out region, defined as $\Omega_c = [0.4, 0.6]^2$, and we wish to solve the multiphysics problem on the rest of domain, denoted as $\Omega_s := \Omega / \Omega_c$. SEM solvers cannot be directly applied because the boundary conditions on the cut-out region are unknown. In this regard, we employ PINNs for the coupled system in $\Omega_c$ to assimilate the data of $\bm{u}$ and $\temp$ while satisfying the PDE defined in \eqref{eq:problem}. In this way, the boundary conditions on $\partial\Omega_c$ can be obtained from evaluating the trained PINNs and their derivatives on $\partial\Omega_c$, and further integrated into the SEM solver.
Specifically in this scenario, we consider $\text{Ra}=10^4$ and assume that we have $N_{\bm{u}}=50$ data of $\bm{u}$, $\{x^{\bm{u}}_i, y^{\bm{u}}_i, \bm{u}_i\}_{i=1}^{N_{\bm{u}}}$, and $N_\temp=50$ data of $\temp$, $\{x_i^\temp, y_i^\temp, \temp_i\}_{i=1}^{N_\temp}$, both of which are randomly sampled from the subdomain $\Omega_c$. We construct the following loss function for PINNs:
\begin{equation}\label{eq:loss_scenario_d}
    \begin{aligned}
        \mathcal{L}(\theta) = & \frac{w_{\bm{u}}}{N_{\bm{u}}}\sum_{i=1}^{N_{\bm{u}}} ||\bm{u}_\theta(x_i^{\bm{u}}, y_i^{\bm{u}}) - \bm{u}_i||_2^2 + \frac{w_{\temp}}{N_\temp}\sum_{i=1}^{N_\temp} |\temp_\theta(x_i^\temp, y_i^\temp) - \temp_i|^2 + \frac{w_{f_1}}{N_f}\sum_{i=1}^{N_f}|(\nabla\cdot \bm{u}_\theta)(x_i^f, y_i^f)|^2 + \\
        & \frac{w_{f_2}}{N_f}\sum_{i=1}^{N_f} ||(\bm{u}_\theta\cdot \nabla\bm{u}_\theta)(x^f_i, y^f_i) + \nabla p_\theta(x^f_i, y^f_i) - \frac{1}{\text{Re}}(\nabla ^2 \bm{u}_\theta)(x^f_i, y^f_i) - \text{Ri}\temp_\theta(x^f_i, y^f_i) \hat{\bm{k}}||_2^2 + \\
        & \frac{w_{f_3}}{N_f}\sum_{i=1}^{N_f}|\bm{u}_\theta(x_i^f, y_i^f)\cdot \nabla \temp_\theta(x^f_i, y^f_i) - \frac{1}{\text{Pe}}\nabla^2\temp_\theta(x^f_i, y^f_i)|^2,
    \end{aligned}
\end{equation}
where $w_{f_1}, w_{f_2}, w_{f_3}$ are hyperparameters to balance different loss terms from Eq. \eqref{eq:problem}, and $x_i^f, y_i^f, i=1,...,N_f$ are residual points. Here, we randomly sample $N_f=1000$ residual points from the subdomain $\Omega_c$. 
We note that in this scenario, we model $\bm{u}, p$ and $\temp$ with NNs, train them simultaneously based on the PINNs loss function \eqref{eq:loss_scenario_d} such that they fit the data and satisfy the physics on the subdomain $\Omega_c$, and no boundary condition of $\bm{u}$ or $\temp$ is imposed in the training. We denote the trained NNs for $\bm{u}$ and $\temp$ as $\bm{u}_{\theta^*}$ and $\temp_{\theta^*}$, respectively. The results of $\bm{u}_{\theta^*}$ and $\temp_{\theta^*}$ are displayed in Fig. \ref{fig:scenario_d_1}, from which we can see that PINNs agree with the reference solutions on $\Omega_c$. 
Subsequently, we evaluate $\bm{u}_{\theta^*}$ and $\temp_{\theta^*}$ at the boundary of $\Omega_c$ to provide the boundary conditions for the SEM solver. In particular, we compute their values to construct the Dirichlet boundary conditions in this study for the SEM solver to be applied on the rest of the domain $\Omega_s$ (the computational domain, the boundary condition, and the mesh are displayed in Fig. \ref{fig:scenario_d_2}). We note, however, that Neumann and Robin conditions can be easily constructed in a similar way due to the differentiability and the mesh-free property of NNs, {\color{cyan}i.e. the value and derivatives of any order of the NN surrogate can be evaluated at any given coordinate.} 
Results from the NeuroSEM are presented in Figs. \ref{fig:scenario_d_3} and \ref{fig:scenario_d_4}. As shown, they agree with the reference solution, demonstrating the effectiveness of our approach in addressing the multiphysics problem \eqref{eq:problem}, where scattered data are only available in a small subdomain.

\subsection{An unsteady-state flow example}\label{sec:3_2}

\begin{figure}[h!]
    \centering
    \includegraphics[trim={0.0cm 0.0cm 0cm 0cm}, clip, width=1.0\textwidth]{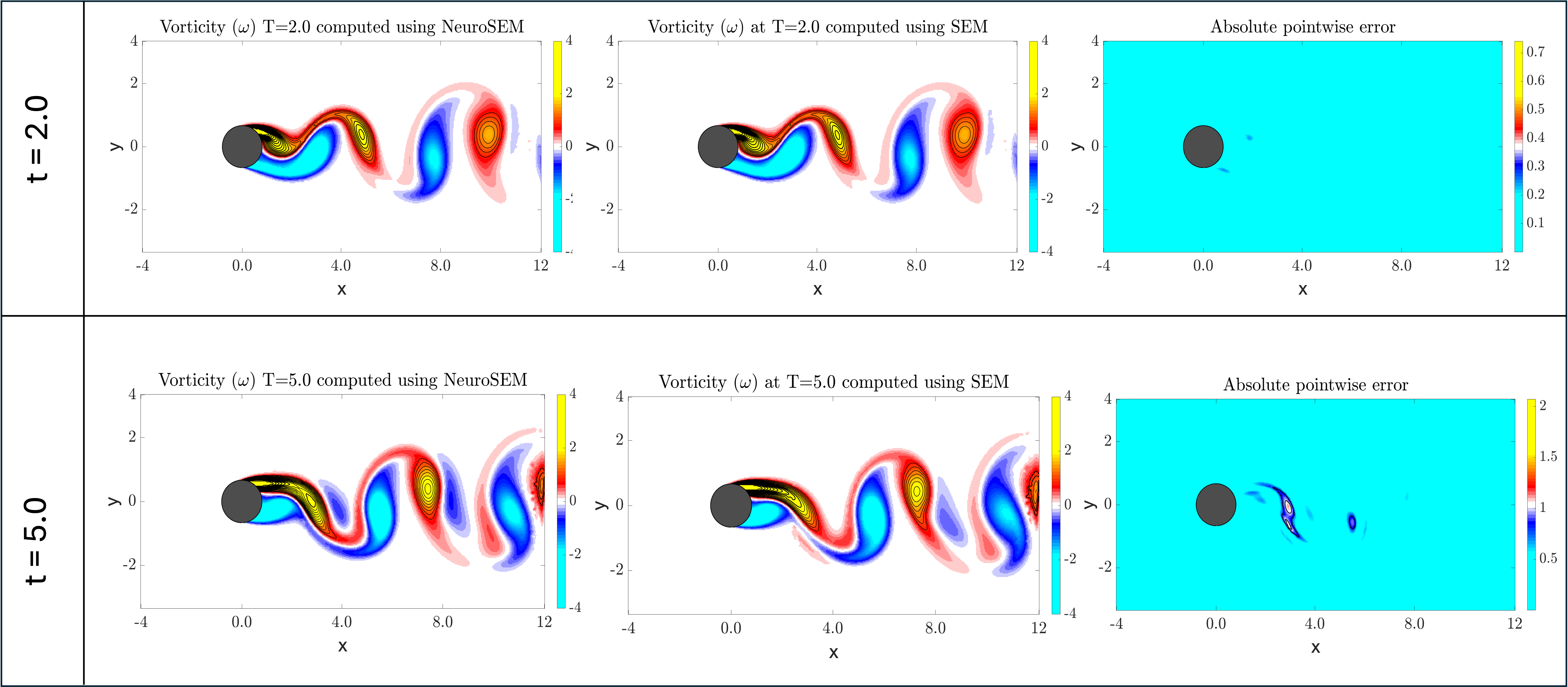}
    \caption{SEM vs NeuroSEM for time-dependent flow past cylinder for $t=2.0$ and $t=5.0$ when $N_t = 21$, where $N_t$ denotes the number of total snapshots used for training (one snapshot per $0.25$ second), from which data of $u, v$ are sampled. We note that the surrogate model of $\temp(x, y, t)$ is trained on $t\in[0, 5]$.}
\label{fig:cyl_1}
\end{figure}


\begin{figure}[h!]
    \centering
    \subfigure[Lift and drag forces]{
    \includegraphics[scale=0.35]{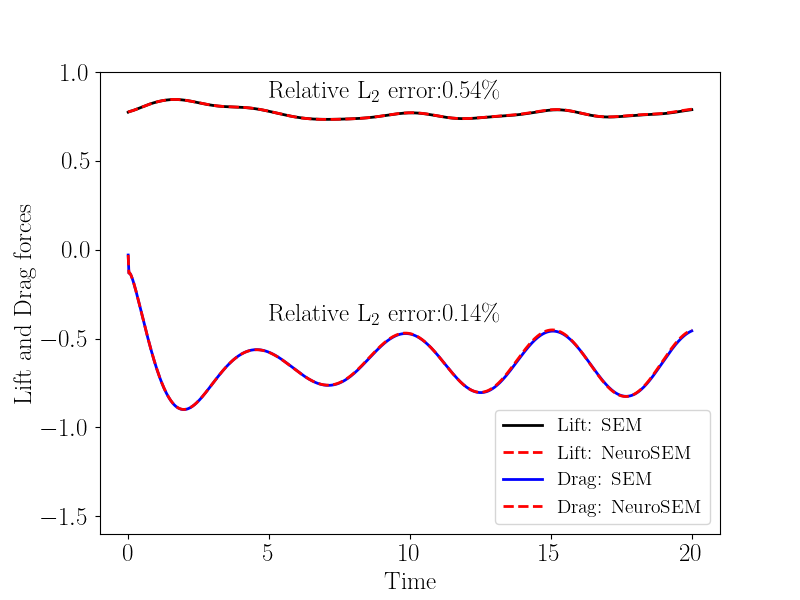}
    }
    \subfigure[Nusselt number]{
    \includegraphics[scale=0.35]{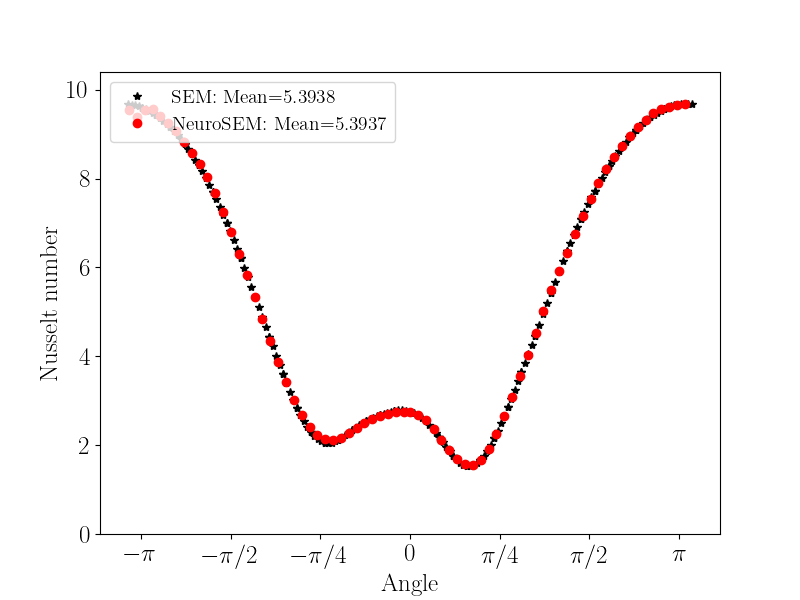}
    }
    \caption{A comparison between (a) lift and drag forces and (b) Local Nusselt number at T=20 computed using NeuroSEM and SEM solvers ($\text{Angle}=\pm\pi$ corresponds to the front stagnation point and $\text{Angle}=0$ corresponds to the rear stagnation point). }
    \label{fig:nusell_ld}
\end{figure}

In this section, we demonstrate the effectiveness of the NeuroSEM method in addressing the multiphysics problem \eqref{eq:problem} for an unsteady flow past a cylinder (domain shown in Fig. \ref{fig:domain}(b)). We set the P\'eclet and the Reynolds number to be $71$ and $100$, respectively, and set $\bm{k} = [0, 1]^T$.
In this example, we assume that we have access to $N_t=21$ snapshots, uniformly distributed on $t\in[0, 5]$ (one snapshot every $\Delta t=0.25$ seconds). In each snapshot, $N_{\bm{u}} = 2,000$ measurements of $\bm{u}$ at random locations are available, denoted as $\{x_i, y_i, t_i, \bm{u}_i\}_{i=1}^{N_{\bm{u}, t}}$ where $N_{\bm{u}, t} = N_{\bm{u}} \times N_t$. We note that in this study we consider the scenario where the locations are different in different snapshots.
We construct the PINNs loss function $\mathcal{L}(\theta)$ as follows:
\begin{equation}\label{eq:loss_flow_past_cylinder}
    \begin{aligned}
        \mathcal{L}(\theta) = & \frac{w_{\bm{u}, t}}{N_{\bm{u}, t}}\sum_{i=1}^{N_{\bm{u}, t}}|\bm{u}_i\cdot \nabla \temp_\theta(x_i, y_i, t_i) - \frac{1}{\text{Pe}}\nabla^2\temp_\theta(x_i, y_i, t_i)|^2 + w_b\mathcal{L}_{b, \temp_\theta}(\theta) + w_i \mathcal{L}_{i, \temp_\theta}(\theta),
    \end{aligned}
\end{equation}
where $\mathcal{L}_{b, \temp_\theta}(\theta)$ and $\mathcal{L}_{i, \temp_\theta}(\theta)$ are the loss functions for the boundary and initial conditions of $\temp$, respectively, and $w_{\bm{u}, t}, w_b, w_i$ are weighting coefficients for the three terms in the PINNs loss function. 
In this scenario, the NN surrogate takes as input $x, y, t$ and outputs $\temp$ and we are in fact solving a time-dependent advection-diffusion system, in which the advection velocity ($\bm{u}$) is also time-dependent, using the PINNs method. 
The trained NN model, denoted as $\temp_{\theta^*}$ where $\theta^*$ denotes the minimizer of the loss function \eqref{eq:loss_scenario_a}, is then fed as a forcing function to the SEM solver to solve the time-dependent NS equations of $\bm{u}$, i.e. Eqs. \eqref{eq:problem_a} and \eqref{eq:problem_b}, with the boundary conditions of $\bm{u}$ (shown in Fig. \ref{fig:domain}(b)). 
We note that the workflow of NeuroSEM in this example is, in fact, the same as the one in \textbf{Scenario A} of the steady-state flow example in Sec. \ref{sec:3_1}, and therefore it can be found in Fig. \ref{fig:case_a}.

Results from NeuroSEM are shown in Fig. \ref{fig:cyl_1}, in which we present the vorticity field, denoted as $\omega$, at $t=2.0$ and $t=5.0$ while the NN surrogate for $\temp$ is trained with snapshots from $t\in[0, 5]$ via the PINNs method. We observe that, compared to the reference, NeuroSEM is able to yield an accurate computation of the vorticity field by solving only the decoupled NS equation, i.e., Eqs. \eqref{eq:problem_a} and \eqref{eq:problem_c}, with the pre-trained PINN as the forcing function. 
In addition, we also compute the lift and drag forces and Nusselt number for a larger time domain $t\in[0, 20]$ (with the same number of measurements per snapshot and one snapshot every $\Delta t= 0.25$ seconds), and display them in Fig. \ref{fig:nusell_ld}(a) and \ref{fig:nusell_ld}(b), respectively. {\color{blue} To generate training data, we ran the SEM solver using a polynomial degree of four, because lower dissipation is required for longer time integration, and the fidelity of the data impacts the convergence of the PINN. The results show excellent agreement for both the forces and the Nusselt number. We also computed the drag and lift using a polynomial degree of 1. The relative \(L_2\) error between lift and drag calculated by SEM and NeuroSEM was 2.39\% and 10.41\%, respectively, and shows a significant divergence at later times.}

{\color{blue}
In a manner similar to the previous example, we also examine a scenario in which the boundary conditions for $T$ are partially missing and noisy data of $T$ from the interior domain are available. Specifically, we assume that the Dirichlet boundary condition on the external boundary is missing and instead we have access to $1,000$ noisy measurements of $T$ in the interior domain for every snapshot (one snapshot every $\Delta t = 0.25$ seconds). In each snapshot, $N_{\bm{u}} = 5,000$ noisy measurements of $\bm{u}$ are available at random locations. The computation is conducted on time domain $t\in[0, 5]$. We also assume that the noise scale for $\bm{u}$ and $T$ in the interior is $0.01$, while the averages of the absolute value of these data before being corrupted by the noise are $0.9598, 0.1296$ for $u, v$ and $0.0763$ for $T$. The relative $L_2$-errors of the velocity magnitude ($\sqrt{u^2 + v^2}$) and the pressure $p$ obtained from NeuroSEM are $2.58\%$ and $7.11\%$, respectively, compared to the reference obtained.
}

{\color{red}
\subsection{NeuroSEM with Particle Imaging Velocimetry (PIV) data}

\begin{figure}[h!]
    \centering
    \subfigure[PIV data.]{
    \includegraphics[scale=0.23]{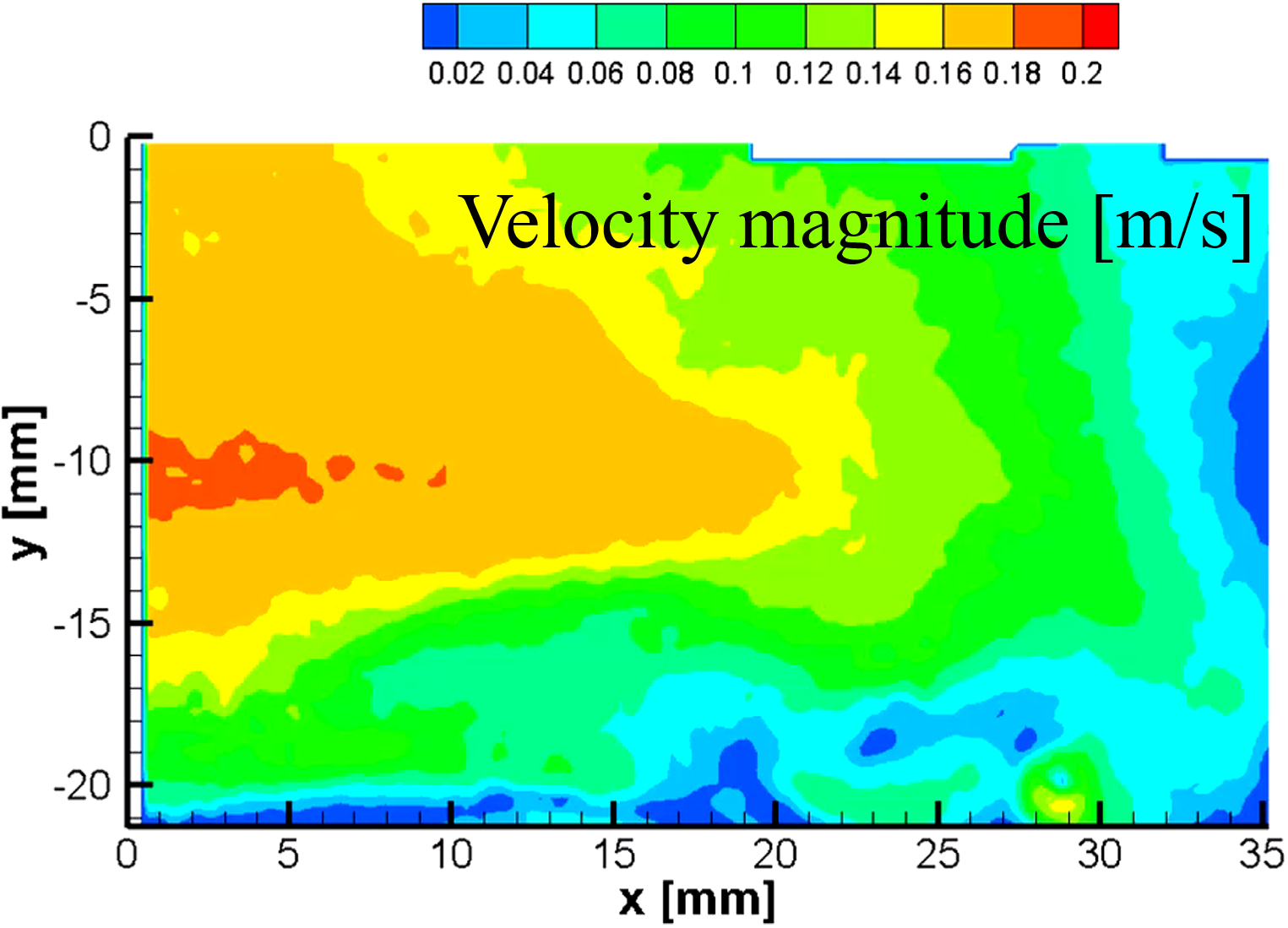}
    }
    \subfigure[PINNs processed PIV data.]{
    \includegraphics[scale=0.23]{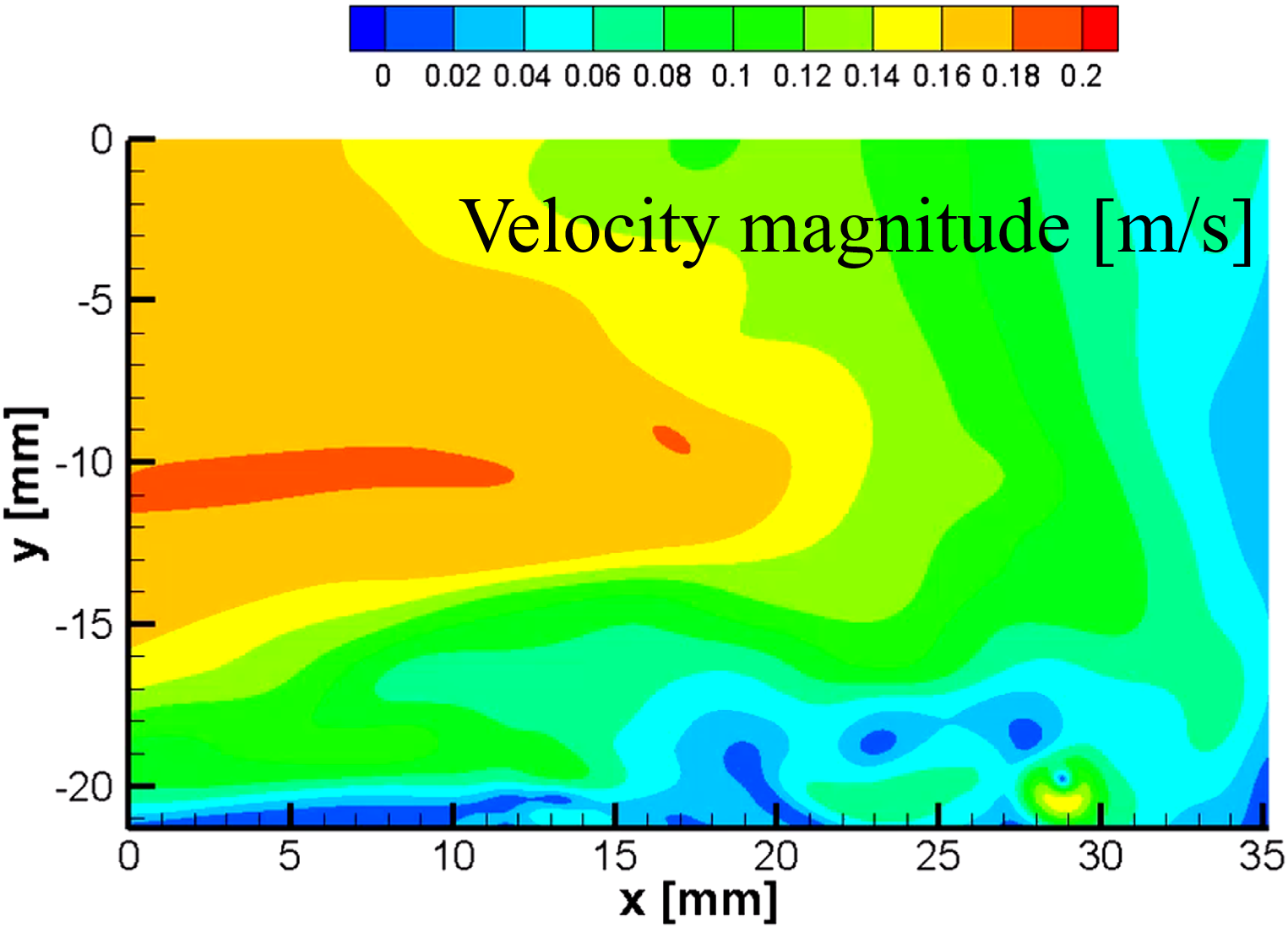}
    }
    \caption{A representative snapshot of the velocity magnitude $(\sqrt{u^2 + v^2})$. (a) The PIV data extracted from particles with orginal fluid domain, and (b) PINNs processed PIV data shown in (a).}
    \label{fig:piv_before_after_pinn}
\end{figure}

We further showcase the presented NeuroSEM method in tackling real PIV data, which are collected for a horseshoe vortex flow in a rectangular domain (as shown in Fig. \ref{fig:piv_before_after_pinn}) $[0, 35] \times [-20, 0]$ (mm). The original PIV dataset has $51$ snapshots of velocity with $\Delta t = 0.001$ and in total $725,423$ data of $\bm{u}$. We consider the following Navier-Stokes equation (Eq. \eqref{eq:problem} with $\text{Ri}=0$) with Reynolds number $\text{Re} = 833.33$:
\begin{subequations}\label{eq:ns}
    \begin{align}
        \nabla \cdot \bm{u} &= 0,\\
        \frac{\partial \bm{u}}{\partial t} + (\bm{u} \cdot \nabla)\bm{u} &= -\nabla p + \frac{1}{\text{Re}} \nabla ^2 \bm{u}.
    \end{align}
\end{subequations}
To simulate this flow with SEM, the initial and boundary conditions are required. Therefore, we first train PINNs to fit the PIV data while satisfying the Navier-Stokes equation thus resulting into a continuous approximation of flowfields $(u,~v,~p)$ in space and time. A representative snapshot of the velocity magnitude computed from PIV and PINNs is shown in Fig. \ref{fig:piv_before_after_pinn}. In addition to the continuous approximation of flow fields, using PINNs instead of regular NNs for processing PIV data offers the advantage of improved regularization. This enhancement is due to the incorporation of physics through the physics-informed loss term in the loss function of PINNs.

To integrate PINNs into the NeuroSEM solver, we present two cases. The first case pertains to the domain illustrated in Fig. \ref{fig:piv_domain}(a), where time-dependent boundary conditions (L, B, R, T in Fig. \ref{fig:piv_domain}[a]) are imposed using PINNs in NeuroSEM framework. The results of NeuroSEM, which utilizes initial conditions derived from a PINN trained on PIV data and the fluid domain (depicted in Fig. \ref{fig:piv_domain}(a)), are shown in Fig. \ref{fig:horseshoe}. In Fig. \ref{fig:horseshoe}(a), we compare the magnitude of the velocity, \(U_{\text{mag}} = \sqrt{u^2 + v^2}\), as obtained from NeuroSEM (left subfigure) and PIV (right subfigure) at a nondimensionalized time \(t = 0.5\). The results show excellent agreement. To further demonstrate the accuracy of the prediction, we computed the vorticity near the wall of the domain, marked by a white dashed box in Fig. \ref{fig:horseshoe}(a). The vorticity contours, which highlight the high vorticity values near the wall corresponding to horseshoe vortices, are clearly observed in both the NeuroSEM and PIV data, as shown in Fig. \ref{fig:horseshoe}(b).

We now conduct a computational experiment for the fluid domain shown in Fig. \ref{fig:piv_domain}. A small cut-out is taken from the original domain, and PINN is used to infer the flow fields within this cut-out. The NeuroSEM solver is run in the subdomain, which is discretized with quadrilateral meshes. At each time step, boundary conditions for \(u\), \(v\), and \(p\) are imposed on \(\partial \Omega_c\) using the PINN. The results of this experiment are presented in Fig. \ref{fig:piv_cutout} at \(t = 4 \times 10^{-2}\). The left subfigure shows \(U_{\text{mag}}\) from NeuroSEM, with a black square indicating the domain used for training. The middle and right subfigures compare \(U_{\text{mag}}\) obtained from NeuroSEM and PIV along the profiles \(P_1\) and \(P_2\), respectively, demonstrating excellent agreement between the results from NeuroSEM and PIV.  

\begin{figure}[h!]
    \centering
    \includegraphics[scale=0.3]{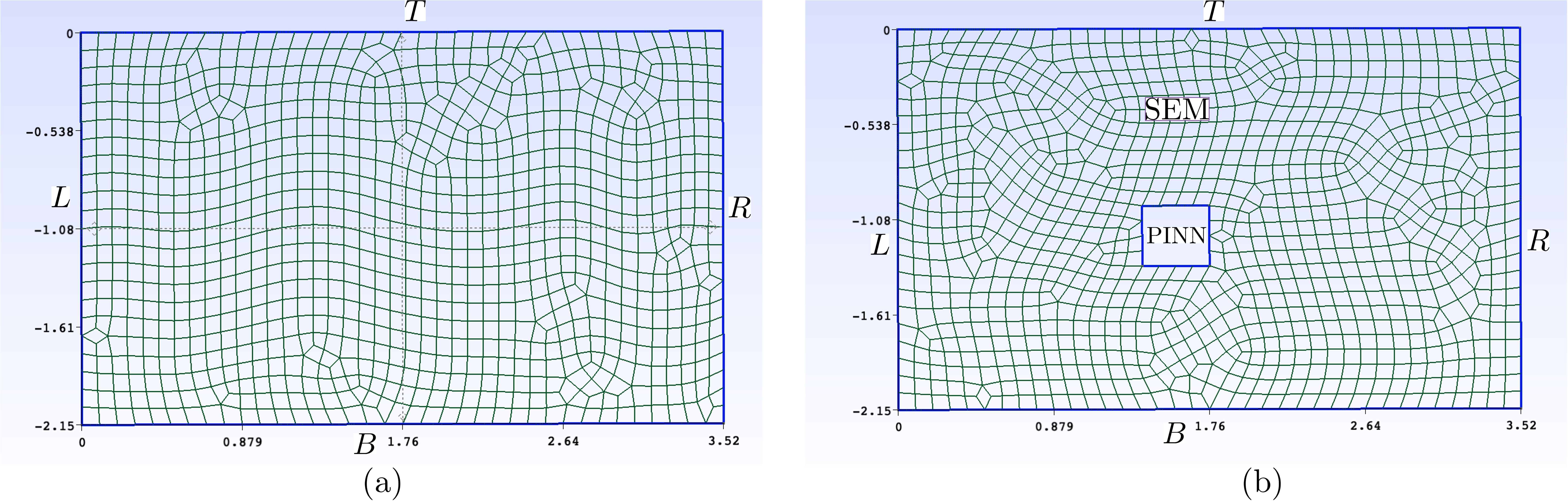}   
    \caption{Domain setup for NeuroSEM to assimilate PIV data. (a) shows the entire domain discretized with quadrilateral meshes, where boundary conditions (L, R, B, T) are imposed by inferring the flow variables in \autoref{eq:ns} using PINN. In (b), the PINN is applied within a subdomain, while \autoref{eq:ns} is solved in the fluid domain, discretized with quadrilateral meshes.}
    \label{fig:piv_domain}
\end{figure}

\begin{figure}[h!]
    \centering
    \subfigure[Magnitude of flow velocity obtained from NeuroSEM and PIV]{
    \includegraphics[width=\textwidth]{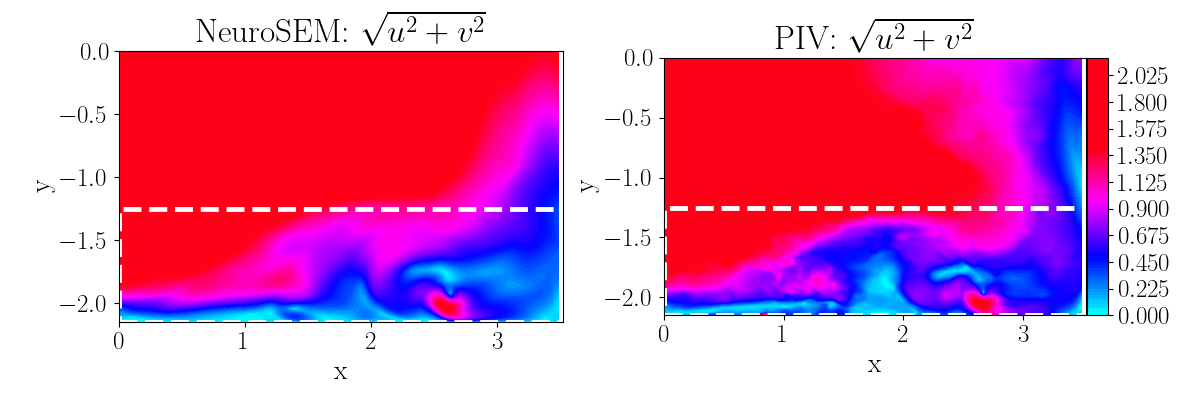}
    }
    \subfigure[Vorticity from NeuroSEM and PIV ]{
    \includegraphics[trim={1cm 0cm 0cm 1cm},clip, width=\textwidth]{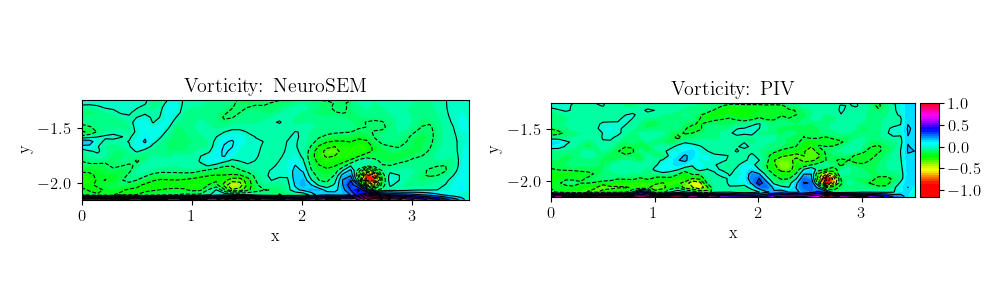}
    }
    \caption{(a) Comparison of the flow velocity magnitude \(\left(\sqrt{u^2 + v^2}\right)\) from NeuroSEM (left) and PIV (right). The white dashed box highlights the formation of the horseshoe vortex. (b) Vorticity plot in the regions outlined by the white dashed box in (a). The high vorticity indicates the presence of the horseshoe vortex near the bottom boundary.}
    \label{fig:horseshoe}
\end{figure}

\begin{figure}[h!]
    \centering
    \includegraphics[width=\textwidth]{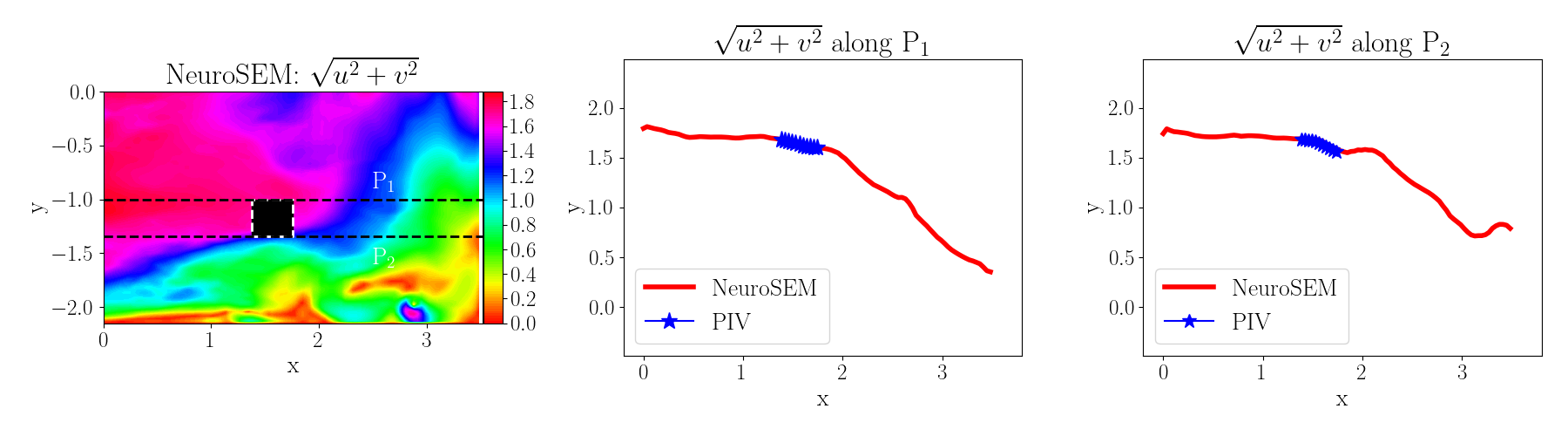}   
    \caption{ Comparison of the flow velocity magnitude \(\left(\sqrt{u^2 + v^2}\right)\) between NeuroSEM and PIV. The left subfigure shows \(U_{\text{mag}}\) from NeuroSEM, with a black square marking the location of the cutout. The middle and right subfigures present the profiles of \(U_{\text{mag}}\) from NeuroSEM, overlaid with \(U_{\text{mag}}\) (blue crosses) obtained from PIV in the cutout region along profiles P1 and P2, respectively.}
    \label{fig:piv_cutout}
\end{figure}
}

\section{Summary}\label{sec:4}
In this work, we present a new computational paradigm by seamlessly coupling deep learning with standard numerical solvers, hence leveraging the relative strengths of both methods. Specifically, we develop a novel NeuroSEM framework, offering a unique algorithmic and software interface for integrating PINNs \cite{raissi2019physics} with the high-fidelity solver Nektar++ \cite{cantwell2015nektar++} for data assimilation purposes. We demonstrate the efficacy of the proposed framework through five scenarios of thermal convection. 
We present results for four scenarios addressing steady-state flow problems, followed by a final scenario that deals with unsteady-state flow conditions. In the first scenario, we reconstruct $[u, v]^\top$ throughout the entire domain using given data on $[u, v]^\top$. In this case, we integrate the PINNs model for $\temp$ in the Navier-Stokes (NS) solver as a forcing function. In the second scenario, we perform a reconstruction of $\temp$ across the entire domain with some measurements of $\temp$. Here, we solve the advection-diffusion equation using the spectral element method by integrating the PINNs model for the given advection velocities. In the third scenario, we have noisy data for \(\bm{u}\) and \(\temp\) at scattered points across the domain, but the boundary conditions for \(\temp\) are not fully specified. Therefore, we develop a PINNs model as a surrogate for \(\temp\) and integrate it into the NS solver as a body force. In the fourth scenario, we address the case where data is provided in a small subdomain. We obtain PINNs surrogates for \(u\), \(v\), \(\temp\), and \(p\) within the subdomain and integrate these models into the SEM solver of Eq. \eqref{eq:problem}, imposing the solutions obtained from the PINNs as appropriate boundary conditions on the boundaries of the subdomain. Finally, we set up a problem for an unsteady flow example describing thermal convection for flow past a cylinder. This setup is similar to the first scenario, except that while integrating the PINNs model into Nektar++, we need to pass an array of time, as the body force in this case is time-dependent. 
{\color{red}To extend the applicability of the proposed NeuroSEM framework, we employed NeuroSEM to integrate the real PIV data for a horseshoe vortex flow.}
Through these five scenarios, we have addressed multiple possible cases that may arise in CFD problems pertaining to data assimilation. Although we focused on Rayleigh-B\'enard Convection, the proposed framework can be seamlessly extended to other physical problems and in three-dimensions. 
Despite considering sufficient data with small noise in the presented study, NeuroSEM can easily address sparse data with large noise by employing certain uncertainty quantification techniques for PINNs \cite{psaros2023uncertainty, zou2024neuraluq, zou2023uncertainty, zou2024leveraging, meng2024hj}.
The NeuroSEM framework integrates the PINNs model through PyTorch's differentiable framework \cite{paszke2019pytorch}, therefore, for large computational domain, inference of the solution from the PINNs model can be easily offloaded to GPUs or any accelerator.

{\color{blue}Despite the capability and flexibility of the PINNs method in handling nonlinearity, physics-informed machine learning and noisy data, its convergence rate across all arbitrary PDEs is still lacking, making it hard to perform convergence analysis for NeuroSEM (unlike conventional data assimilation methods such as 3D-VAR, 4D-VAR and Kalman filters \cite{law2015data, fletcher2022data, evensen2022data}. The error of NeuroSEM will be bounded from below by $\mathcal{O}(h^{n+1})$, where $h$ is the length of the edge of a mesh element in the SEM step and $n$ the polynomial order, and from above by the error from PINNs:
$$
\mathcal{O}(h^{n+1}) \le \mathcal{E} \le \mathcal{O}(\mathcal{L}_{PINN}),
$$
where $(\mathcal{L}_{PINN})$ represents the total loss of PINNs.}
{\color{cyan}Another limitation of the presented approach is that successful training of PINNs, which often requires sufficient data \cite{raissi2019physics, jin2021nsfnets, cai2021physics, cai2021physics_fluid, zou2023correcting}, is essential for NeuroSEM to perform accurate computations.} {\color{green}For time-dependent problems, such as those considered in this work, PINNs must be well-trained across the entire time domain to serve effectively as a surrogate model.}


\section*{Acknowledgements}
We acknowledge the support of the MURI/AFOSR FA9550-20-1-0358 project and the
DOE-MMICS SEA-CROGS DE-SC0023191 award. G.E.K. is supported by the ONR Vannevar
Bush Faculty Fellowship (N00014-22-1-2795). 
We also thank Dr. Chris Cantwell from Imperial College, London and Dr. Ann Almgren from Lawrence Berkeley National Laboratory for insightful discussions, and Prof. Shengze Cai from Zhejiang University for his help with the horseshow vortex flow and PIV data.

\newpage

\bibliographystyle{unsrt}
\bibliography{ref}

\appendix
\newpage

\section{Session file for the integration of PINNs into Nektar++}\label{appendix_session}

In this section, we present details for the integration of physics-informed neural networks (PINNs) \cite{raissi2019physics} into solvers based on spectral element methods (SEM) for solving the multiphysics problem \eqref{eq:problem} in different cases. Specifically, we present the Session file (session.xml) for Nektar++ \cite{cantwell2015nektar++} for following three cases:
\begin{enumerate}
    \item Case (A): Given a neural network (NN) surrogate of \( T \) trained using the PINN method, we solve Eqs. \eqref{eq:problem_a} and \eqref{eq:problem_b} using Nektar++. The session file for Nektar++ is provided in \autoref{code:phi_PINN}, where PINN model is parsed as bodyforcing function.
    \item Case (B): Given a NN surrogate of $\bm{u}$, trained using the PINN method, we solve Eq. \eqref{eq:problem_c} using Nektar++. The session file for Nektar++ is provided in \autoref{code:uv_PINN}, where PINN model is parsed to compute the advection velocities.
    \item Case (C): Given the PINN model trained in a subdomain, we integrate it with Nektar++ by inferring the boundary values of \([u, v]^T,~p\), and \(\temp\) from the PINN and imposing them as boundary conditions to solve Eq. \eqref{eq:problem}. The session file for Nektar++ to solve the problem is shown in \autoref{code:cutout_PINN}.
\end{enumerate}

\lstset{
    escapeinside=``,
    language=xml,
    tabsize=3,
    caption=Session file (session.xml) for Nektar++ for integrating PINN for $\temp \rightarrow (u~v)$,
    label=code:phi_PINN,
    basicstyle=\ttfamily{pcr}\footnotesize,
    frame=shadowbox,
    rulesepcolor=\color{gray},
    xleftmargin=14pt,
    framexleftmargin=14pt,
    keywordstyle=\color{blue}\bf,
    commentstyle=\color{green},
    stringstyle=\color{red},
    numbers=left,
    numberstyle=\tiny,
    numbersep=5pt,
    breaklines=true,
    showstringspaces=false,
    basicstyle=\footnotesize,
    emph={FUNCTION,FORCE, BODYFORCE},emphstyle={\color{magenta}}}
\begin{lstlisting}
<?xml version="1.0" encoding="utf-8"?>
<NEKTAR>
    <EXPANSIONS>
        <E COMPOSITE="C[9]" NUMMODES="2" TYPE="MODIFIED" FIELDS="u, v, p"/>      
    </EXPANSIONS>
    <CONDITIONS>
       <PARAMETERS>
            <P> FineTime       = 100                    </P>
            <P> TimeStep       =  0.01                  </P>
            <P> NumSteps       = FinTime/TimeStep       </P>
            <P> IO_CheckSteps  =  1000                  </P>
            <P> IO_InfoSteps   =  100                   </P>
            <P> Re             =  118.68                </P>
            <P> Kinvis         = 1/Re                   </P>
        </PARAMETERS>
        <FUNCTION NAME="PINNBodyForce">
            <E VAR="u" VALUE="PINN.pt" />
            <E VAR="v" VALUE="PINN.pt" />
        </FUNCTION>
    </CONDITIONS>
    <FORCING>
        <FORCE TYPE="Body">
            <BODYFORCE> BodyForce </BODYFORCE>
        </FORCE>
    </FORCING>
</NEKTAR>
\end{lstlisting}

\lstset{
    float=!ht,
    language=xml,
    tabsize=3,
    caption=Session file (session.xml) for Nektar++ for integrating PINN for $(u~v) \rightarrow \temp$,
    label=code:uv_PINN,
    frame=shadowbox,
    rulesepcolor=\color{gray},
    xleftmargin=14pt,
    framexleftmargin=14pt,
    keywordstyle=\color{blue}\bf,
    commentstyle=\color{green},
    stringstyle=\color{red},
    numbers=left,
    numberstyle=\tiny,
    numbersep=5pt,
    breaklines=true,
    showstringspaces=false,
    basicstyle=\footnotesize,
    emph={FUNCTION,FORCE, BODYFORCE},emphstyle={\color{magenta}}}
\begin{lstlisting}
<?xml version="1.0" encoding="utf-8"?>
<NEKTAR>
    <EXPANSIONS>
        <E COMPOSITE="C[9]" NUMMODES="2" TYPE="MODIFIED" FIELDS="u" />      
    </EXPANSIONS>
    <CONDITIONS>
       <PARAMETERS>
            <P> FineTime       = 1000                             </P>
            <P> TimeStep       =  0.01                            </P>
            <P> NumSteps       = FinTime/TimeStep                 </P>
            <P> IO_CheckSteps  =  1000                            </P>
            <P> IO_InfoSteps   =  100                             </P>
            <P> epsilon             =  0.011867816581938534       </P>
            <P> ax                  = "PINN"                       </P>
            <P> ay                  = "PINN"                       </P>
        </PARAMETERS>
	   <FUNCTION NAME="PINNAdvectionVelocity">
            <E VAR="Vx" VALUE="PINN.pt" />
            <E VAR="Vy" VALUE="PINN.pt" />
        </FUNCTION>
     </CONDITIONS>
</NEKTAR>
\end{lstlisting}

\lstset{
    float=!ht,
    language=xml,
    tabsize=3,
    caption=Session file (session.xml) of Nektar++ integration of PINN model for the boundary conditions,
    label=code:cutout_PINN,
    frame=shadowbox,
    rulesepcolor=\color{gray},
    xleftmargin=14pt,
    framexleftmargin=14pt,
    keywordstyle=\color{blue}\bf,
    commentstyle=\color{green},
    stringstyle=\color{red},
    numbers=left,
    numberstyle=\tiny,
    numbersep=5pt,
    breaklines=true,
    showstringspaces=false,
    basicstyle=\footnotesize,
    emph={FUNCTION,FORCE, BODYFORCE},emphstyle={\color{magenta}}}
\begin{lstlisting}
<?xml version="1.0" encoding="utf-8" ?>
<NEKTAR>
    <EXPANSIONS>
        <E COMPOSITE="C[17]" NUMMODES="3" TYPE="MODIFIED" FIELDS="u,v,theta,p" />
    </EXPANSIONS>
    <CONDITIONS>
       <PARAMETERS>
            <P> TimeStep       = 0.1             </P>
            <P> TFinal         = 400            </P>
            <P> NumSteps       = TFinal/TimeStep </P>
            <P> IO_CheckSteps  = 200/TimeStep    </P>
            <P> IO_InfoSteps   = 1               </P>
            <P> IO_CFLSteps    = 1       	     </P>           
            <P> Ra            = 10000             </P>
            <P> Pr            = 0.71              </P> 
            <P> Kinvis        = (Pr/Ra)^0.5       </P>
        </PARAMETERS>
    
        <BOUNDARYREGIONS>
            <B ID="0"> C[9]        </B>   <!-- left_big   -->
            <B ID="1"> C[10]  </B>   <!-- right_big  -->
            <B ID="2"> C[11]  </B>   <!-- top_big    -->
            <B ID="3"> C[12]  </B>   <!-- bot_big    -->
            <B ID="4"> C[13]  </B>   <!-- bot_small  -->
            <B ID="5"> C[14]  </B>   <!-- right_small-->
            <B ID="6"> C[15]  </B>   <!-- top_small  -->
            <B ID="7"> C[16]  </B>   <!-- left_small -->
        </BOUNDARYREGIONS>
        <BOUNDARYCONDITIONS>
            <REGION REF="0">
                <D VAR="u"      VALUE="0" />
                <D VAR="v"      VALUE="0" />
                <N VAR="theta"  VALUE="0" />
                <N VAR="p"      VALUE="0" USERDEFINEDTYPE="H" />
            </REGION>
            <REGION REF="1">
                <D VAR="u"      VALUE="0" />
                <D VAR="v"      VALUE="0" />
                <N VAR="theta"  VALUE="0" />
                <N VAR="p"      VALUE="0" USERDEFINEDTYPE="H" />
            </REGION>
            <REGION REF="2">
                <D VAR="u"      VALUE="0" />
                <D VAR="v"      VALUE="0" />
                <D VAR="theta"  VALUE="-0.5" />
                <N VAR="p"      VALUE="0" USERDEFINEDTYPE="H" />
            </REGION>
            <REGION REF="3">
                <D VAR="u"      VALUE="0" />
                <D VAR="v"      VALUE="0" />
                <D VAR="theta"  VALUE="0.5" />
                <N VAR="p"      VALUE="0" USERDEFINEDTYPE="H" />
            </REGION>
            <!-- bot small -->
            <REGION REF="4">
                <D VAR="u"      FILE="RBCavity_bot_small_u.fld" />
                <D VAR="v"      FILE="RBCavity_bot_small_v.fld" />
                <D VAR="theta"  FILE="RBCavity_bot_small_theta.fld" />
                <D VAR="p"      FILE="RBCavity_bot_small_p.fld" />
            </REGION>
            <!-- right small -->
            <REGION REF="5">
                <D VAR="u"      FILE="RBCavity_right_small_u.fld" />
                <D VAR="v"      FILE="RBCavity_right_small_v.fld" />
                <D VAR="theta"  FILE="RBCavity_right_small_theta.fld" />
                <D VAR="p"      FILE="RBCavity_right_small_p.fld" />
            </REGION>
            <!-- top small -->
            <REGION REF="6">
                <D VAR="u"      FILE="RBCavity_top_small_u.fld" />
                <D VAR="v"      FILE="RBCavity_top_small_v.fld" />
                <D VAR="theta"  FILE="RBCavity_top_small_theta.fld" />
                <D VAR="p"      FILE="RBCavity_top_small_p.fld" />
            </REGION>
            <!-- left small -->
            <REGION REF="7">
                <D VAR="u"      FILE="RBCavity_left_small_u.fld" />
                <D VAR="v"      FILE="RBCavity_left_small_v.fld" />
                <D VAR="theta"  FILE="RBCavity_left_small_theta.fld" />
                <D VAR="p"      FILE="RBCavity_left_small_p.fld" />
            </REGION>
        </BOUNDARYCONDITIONS>
    </CONDITIONS>
</NEKTAR>
\end{lstlisting}

\section{Details of PINNs}\label{sec:appendix_pinn} 

In this section, we present details of PINNs in the numerical experiments, including the NN architecture and the training procedure.

\subsection{The steady-state flow example}

Recall that in the steady-state flow example we considered four scenarios. In \textbf{Scenario A}, we employed a NN with four hidden layers, each of which has $100$ neurons and is equipped with hyperbolic tangent activation function. The NN takes as input the spatial coordinates $x, y$, and outputs the temperature $\temp$.
In \textbf{Scenario B}, we employed a NN with the same number of hidden layers and the same widths and activation functions of these hidden layers, and . The NN takes as input the spatial coordinate $x, y$ and outputs $u, v$ and $p$.
In both scenarios, Adam optimizer \cite{kingma2014adam} was employed for $100,000$ iterations with the following piecewise constant learning rate scheduler for stable training of PINNs \cite{wang2023solution, jin2021nsfnets, zou2023correcting} for $\text{Ra}=10^4, 10^5$: $1\times10^{-3}$ for the first $50,000$ iterations and $1\times10^{-4}$ for another $50,000$ iterations.
For $\text{Ra}=10^6$, Adam optimizer was employed for $300,000$ iterations with the following piecewise constant learning rate scheduler: $1\times 10^{-3}$ for the first $200,000$ iterations and $1\times 10^{-4}$ for another $100,000$ iterations.
In \textbf{Scenario C}, we only considered $\text{Ra}=10^4$ and the NN architecture and the training procedure were the same as the ones in \textbf{Scenario A} with $\text{Ra}=10^4$. 
In \textbf{Scenario D}, we employed two NNs, which have four hidden layers ($100$ neurons and hyperbolic tangent activation function) and output $u, v, p$ and $\temp$, respectively. Adam optimizer was employed for $600,000$ iterations with the following piecewise constant learning rate scheduler:
$1\times10^{-3}$ for $200,000$ iterations, $5\times10^{-4}$ for another $200,000$ iterations, and $1\times10^{-4}$ for the last $200,000$ iterations. {\color{cyan}For all scenarios of this example, the weighting coefficients in the PINN loss function are set to one.}

\subsection{The unsteady-state flow example}

In the flow past cylinder example, we employed a NN to model the time-dependent behavior of $\temp$. The NN has four hidden layers ($100$ neurons and hyperbolic tangent activation function for each layer) and takes as input $x, y, t$ and output $\temp$. Adam optimizer was employed for $1,000,000$ iterations with a piecewise constant learning rate scheduler: $1\times 10^{-3}$ for $500,000$ iterations, $5\times10^{-4}$ for another $200,000$ iterations, $1\times10^{-4}$ for the following $200,000$ iterations, and $1\times 10^{-5}$ for the last $100,000$ iterations. {\color{cyan}In this example, the weighting coefficients in the PINN loss function are set to one.}

{\color{red}
\subsection{The real PIV data example}

In this example, we employed a NN to model the time-dependent behavior of $\bm{u}$ and $p$. The NN has five hidden layers ($100$ neurons and hyperbolic tangent activation function for each layer) and takes as input $x, y, t$ and output $u, v, p$. $1,000,000$ residual points are randomly sampled across the spatial-temporal domain according the Latin hypercube sampling method. Adam optimizer was employed for $1,000$ epochs with learning rate $1\times 10^{-3}$. The mini-batch training strategy was adopted and the batch size was chosen as $50,000$. {\color{cyan}The weight coefficient for the data loss term is set to $10$ while the weight coefficient for the PDE loss term is set to one.}
}





\section{Acceleration of training of PINNs}\label{accl_pinn}
In this section, we build up on the vanilla implementation of the PINNs method for the steady-state flow example and explore other variations of PINNs such as separable PINNs \cite{cho2023separable}, gradient-enhanced PINNs \cite{Yu_2022} and self-adaptive PINNs \cite{mcclenny2020self, MCCLENNY2023111722} as well as their different combinations for $\text{Ra} = 10^4$ and $\text{Ra} = 10^5$ to observe the accuracy and accelerate the training process. For this section, we implemented the vanilla PINN with five layers, $32$ neurons in each of these layers and $\tanh$ activation function. Adam\cite{kingma2014adam} optimizer with a learning rate of $10^{-4}$ was employed for $600,000$ iterations for both values of the Rayleigh number. 
For the case with $\text{Ra}=10^5$, the NN parameter was initialized as the one obtained from the $\text{Ra}=10^4$ case. A systematic evaluation of each model's speed and accuracy is given in Table \ref{tab:acceleration_ablationstudy_104} for $\text{Ra} = 10^4$ and Table \ref{tab:acceleration_ablationstudy_105} for $\text{Ra} = 10^5$. The following subsections discuss the implementation details of each of these methods.

\subsection{Using separable PINNs}
In separable PINNs \cite{cho2023separable}, instead of feeding a stacked matrix of $x,y$ coordinates together into the NN, we have separated sub-networks, which take on each independent one-dimensional coordinate as input and generate a final output by an aggregation module involving a simple outer product and summation. 
Before feeding the inputs into the sub-networks, there is a step where we ensure that each input has at least two dimensions, so that the dense layer can process it properly.
Our implementation is detailed in Table \ref{SPINNImplementation} for $\text{Ra} = 10^4$ and $\text{Ra} = 10^5$. Moreover, we 
observed that the convergence, in this case, is sensitive to the choice of the seed, and also the training time is sensitive to the kind of relative error printed at regular intervals (e.g., every $100$ epochs). 

\begin{table}[!h]
    \centering
    \begin{tabular}{c|c} \hline \hline
         Number of collocation points for $\text{Ra} = 10^4$ & $100 \times 100$\\\hline
         Number of collocation points for $\text{Ra} = 10^5$ & $1000 \times 1000$\\\hline
         Number of layers for $\text{Ra} = 10^4$ & 5\\ \hline
         Number of layers for $\text{Ra} = 10^5$ & 6\\ \hline
         Neurons per layer& 32\\ \hline 
         Activation function& $\tanh$\\ \hline 
         Learning rate& $10^{-4}$\\ \hline 
 Value of the weighting coefficients &1.0\\\hline
 Iterations& $600000$\\\hline 
 ML Framework& JAX-Flax \cite{flax2020github}\\ \hline 
 GPU& GeForce RTX 3090\\ \hline \hline
    \end{tabular}
    \caption{Separable PINNs implementation details.}
    \label{SPINNImplementation}
\end{table}

\subsection{Using self-adaptive PINNs}
In the vanilla implementation of the PINNs method, each term in the loss function is weighted heuristically. In this section, our implementation of the self-adaptive method \cite{mcclenny2020self, MCCLENNY2023111722}, in which the weighting coefficients are automatically computed based on the residual value by penalizing more the high residual value, is such that we introduced flexible adaptability to the weighting coefficients corresponding to the residual and boundary loss terms, denoted by $w_{\bm{u}}$ and $w_{\bm{b}}$, respectively, and reformulated our loss function as following:
\begin{equation}
        \mathcal{L}(\theta) = \mathcal{L}_{\bm{b}} (\theta, w_{\bm{b}}) + \mathcal{L}_{\bm{u}}(\theta, w_{\bm{u}}) 
\end{equation}
Each loss term can be expanded, taking into consideration that we have both Dirichlet and Neumann boundary terms, as follows:
\begin{multline}
\mathcal{L}(\theta) = \frac{1}{N_b} \sum_{i=1}^{N_b} (w_b^i \: \nabla_x T_{\theta_{NN}}^{i})_{lb}^2 + \frac{1}{N_b} \sum_{i=1}^{N_b} (w_b^i\: \nabla_x T_{\theta_{NN}}^i)_{rb}^2 + \frac{1}{N_b} \sum_{i=1}^{N_b} \left(w_b^i (T_{\theta_{NN}}^i - T_{\theta_{tb}}^i) \right)^2 +\\ \frac{1}{N_b} \sum_{i=1}^{N_b} \left( w_b^i (T_{\theta_{NN}}^i - T_{\theta_{bb}}^i) \right)^2 + \frac{1}{N_u} \sum_{i=1}^{N_u} (w_u^i  f^{i})^2 
\end{multline}
where $tb,bb$ correspond to top and bottom boundaries respectively and $lb,rb$ correspond to the left and right boundaries respectively, $f^i$ corresponds to the residual term given in equation \ref{eq:loss_scenario_a} which is supposed to be $0$ at each collocation point $i$ and each $w_{\bm{j}} = (w_{\bm{j}}^1, w_{\bm{j}}^2, \cdots, w_{\bm{j}}^{N_i})$ are trainable, self-adaptive weighting coefficients, $j \in \{b,u\}$ such that $N_{\bm{b}}$ is the number of boundary points, $N_{\bm{u}}$ is the number of collocation points. 
In our implementation, we performed gradient ascent on the loss function $\mathcal{L}$ with respect to $w_{\bm{i}}$ for $i \in \{b,u\}$.
Table \ref{SAImplementation} highlights the implementation details. It is also important to note that the training for this part is done via serialization techniques \cite{kidger2021equinox} to save and restore the state of the model and training parameters after every $200,000$ iterations, until $600,000$ iterations were reached. 
The time was measured by summing up the training times of each of the three parts but the error was measured once the training for all $600,000$ iterations was completed. 

\begin{table}[!h]
    \centering
    \begin{tabular}{c|c} \hline \hline
         Number of collocation points& $10000$\\ \hline 
         Number of boundary points& $1000$\\ \hline 
         Number of layers& 5\\ \hline 
         Neurons per layer & 32\\ \hline 
         Activation function& $\tanh$\\ \hline 
         Learning rate& $10^{-4}$\\ \hline 
         Value of the weighting coefficients & adaptive; randomly initialized from $U[0.01,1]$ \\ \hline 
         Iterations& $600000$  \\ \hline 
         ML Framework & JAX-Equinox \cite{kidger2021equinox}\\ \hline 
         GPU& GeForce RTX 3090\\ \hline \hline
    \end{tabular}
    \caption{Self-adaptive PINNs implementation details.}
    \label{SAImplementation}
\end{table}

\subsection{Using gradient-enhanced PINNs}
In gradient-enhanced PINNs \cite{Yu_2022}, an additional loss term to the equation \eqref{eq:loss_scenario_a} is introduced to enforce the gradient of the PDE residual to be $0$ at some $N_g$ points. Denoting the weighting coefficient by $w_{\bm{g}}$, this loss function term is expressed as:
\begin{equation}\label{eq:gpinnloss}
    \begin{aligned}
        \mathcal{L}_{\bm{g}}(\theta) = & \frac{w_{\bm{g}}}{N_{\bm{g}}}\sum_{i=1}^{N_{\bm{g}}}|\bm{u}_i\cdot \nabla^2 \temp_\theta(x_i, y_i) - \frac{1}{\text{Pe}}\nabla^3\temp_\theta(x_i, y_i)|^2,
    \end{aligned}
\end{equation}
Corresponding to the two input dimensions $x,y$, we have gradient of residuals computed with respect to both $x,y$ and incorporated in our final loss function.
For $\text{Ra} = 10^4$, we fixed $w_{\bm{g}} = 10^{-4}$ and for $\text{Ra} = 10^5$, we fixed $w_{\bm{g}} = 10^{-6}$. A detailed information of our implementation is given in table \ref{gPINNImplementation}. 
\begin{table}[!h]
    \centering
    \begin{tabular}{c|c} \hline \hline
         Number of collocation points& 10000\\ \hline 
         Number of boundary points& 1000\\\hline
 Number of gradient calculation points&100\\\hline 
         Number of layers& 5\\ \hline 
         Neurons per layer& 32\\ \hline 
         Activation function& $\tanh$\\ \hline 
         Learning rate& $10^{-4}$\\ \hline 
         Weighting coefficients for $Ra = 10^4$ & $w_{\bm{b}} = 1.0, w_{\bm{u}} = 1.0, w_{\bm{g}} = 10^{-4}$\\ \hline 
         Weighting coefficients for $Ra = 10^5$ & $w_{\bm{b}} = 1.0, w_{\bm{u}} = 1.0, w_{\bm{g}} = 10^{-6}$\\ \hline 
         Iterations& $600000$\\ \hline 
         ML Framework& JAX-Equinox \cite{kidger2021equinox}\\ \hline 
         GPU& GeForce RTX 3090\\ \hline \hline
    \end{tabular}
    \caption{Gradient-enhanced PINNs implementation details.}
    \label{gPINNImplementation}
\end{table}

\subsection{Implementation details about combinations of aforementioned variations of the PINNs class performed for $\text{Ra} = 10^4$:}
\begin{itemize}
    \item \emph{Separable, self-adaptive PINNs}: In our implementation of separable PINNs, we introduced self-adaptive weighting coefficients that are initialized from a uniform distribution between $0.01- 1.00$, and trained with a learning rate of $10^{-4}$, Adam optimizer \cite{kingma2014adam} for $600,000$ iterations. 

    \item \emph{Gradient-enhanced, self-adaptive PINNs}: In our implementation of gradient-enhanced PINNs, we introduced self-adaptive weighting coefficients, all initialized from a uniform distribution, and we further multiplied the initialized value of $w_{\bm{g}}$, which is from a uniform distribution between $0.01- 1.00$ by $10^{-4}$, and trained with a learning rate of $10^{-4}$ and Adam \cite{kingma2014adam} optimizer for $600,000$ iterations. The gradient of the residual was calculated only on $100$ collocation points. 

    \item \emph{Gradient-enhanced, separable PINNs}: In our implementation of separable PINNs, we introduced an additional loss term to enforce the gradient of the PDE residual to be $0$ on $100 \times 100$ collocation points. We trained the model with a learning rate of $10^{-4}$ and Adam \cite{kingma2014adam} optimizer for $600,000$ iterations, and the weighting coefficients of the data and residual loss terms were set to $1.0$, with $w_{\bm{g}} = 10^{-4}$.

\item \emph{Separable, gradient-enhanced, self-adaptive PINNs}: Combining the aforementioned three approaches, we implemented this model with Adam \cite{kingma2014adam} optimizer, a learning rate of $10^{-4}$ and trained for $600,000$ iterations. 
\end{itemize}

\begin{table}
    \centering
   {
    \begin{tabular}{p{6cm}|c|c} \hline \hline
         \textbf{Method}&  \textbf{Avg. Training Time} (sec)&  \textbf{Relative $L_2$-error} $(\%)$\\ \hline 
          Vanilla PINNs & 1481.96 &0.19\\\hline
         Separable PINNs &  769.03 &  5.82\\ \hline 
         Self-adaptive PINNs &  4023.30 &  0.19\\ \hline 
         Separable, self-adaptive PINNs & 856.73 &  7.02\\ \hline 
         Gradient-enhanced PINNs &  2194.03&  0.52\\ \hline  
 Gradient-enhanced, self-adaptive PINNs & 6118.57 & 1.31 \\ \hline
        Separable, gradient-enhanced PINNs & 1404.67 &  2.16\\\hline
 Separable, gradient-enhanced, self-adaptive PINNs & 1601.98 & 7.67\\\hline \hline
    \end{tabular}}
    \caption{Average training time and relative $L_2$-error for various models corresponding to $Ra = 10^4$ over three runs on GeForce RTX 3090 GPU. Note that due to compilation differences, training times may vary by approximately $30$ to a few hundred seconds. Note that due to difference in the GPU nodes, slightly different error values may be observed.}
    \label{tab:acceleration_ablationstudy_104}
\end{table}

\begin{table}
    \centering
   {
    \begin{tabular}{c|c|c} \hline \hline
         \textbf{Method}&  \textbf{Avg. Training Time} (sec)&  \textbf{Relative $L_2$-error} $(\%)$\\ \hline 
           \resizebox{!}{8pt}{Vanilla PINNs} & 1487.18 & 0.52\\\hline
           \resizebox{!}{8pt} {Gradient-enhanced PINNs} & 2233.48 & 1.02 \\ \hline
          \resizebox{!}{8pt}{Separable PINNs} &  1889.35 &  2.23\\ \hline  
          \resizebox{!}{8pt}{Self-adaptive PINNs} &  4000.69 &  0.27\\ \hline \hline
    \end{tabular}}
    \caption{Average training time and relative $L_2$-error for various models corresponding to $Ra=10^5$ over three runs on GeForce RTX 3090 GPU. Note that due to compilation differences, training times may vary by approximately $30$ to a few hundred seconds. Note that due to difference in the GPU nodes, slightly different error values may be observed.}
    \label{tab:acceleration_ablationstudy_105}
\end{table}

\end{document}